\documentclass[table]{article}

\usepackage{PRIMEarxiv}

\usepackage[utf8]{inputenc}
\usepackage[T1]{fontenc}
\usepackage{url}
\usepackage{nicefrac}
\usepackage{microtype}
\usepackage{lipsum}
\usepackage{wrapfig}
\usepackage{enumitem}
\usepackage{fancyhdr}

\usepackage{amsmath}
\usepackage{amssymb}
\usepackage{amsfonts}

\usepackage{colortbl}
\usepackage{multirow}
\usepackage{siunitx}
\usepackage{booktabs}
\usepackage{caption}
\usepackage{array}
\usepackage{rotating}

\usepackage{graphicx}
\graphicspath{{media/}}
\usepackage{subcaption}
\usepackage{pgfplots}
\pgfplotsset{compat=1.18}
\usepackage{tikz}
\usetikzlibrary{positioning,fit,shapes.geometric,shadows,arrows.meta}
\usepackage{tcolorbox}
\tcbuselibrary{breakable, skins}
\usepackage{listings} 

\usepackage[noend]{algpseudocode}
\usepackage{algorithm}

\usepackage{hyperref}
\usepackage{cleveref}
\usepackage[numbers]{natbib}
\usepackage{placeins}

\usepackage{authblk}
\usepackage{academicons}
\usepackage{orcidlink}
\usepackage{fontawesome5}

\newcommand{\nmetrics}{n_{\mathrm{metrics}}}

\definecolor{lightyellow}{rgb}{1, 1, 0.88}
\definecolor{highlightcolor}{rgb}{1,1,0.7}
\definecolor{codegreen}{rgb}{0,0.6,0}
\definecolor{codegray}{rgb}{0.5,0.5,0.5}
\definecolor{codepurple}{rgb}{0.58,0,0.82}
\definecolor{backcolour}{gray}{0.97}
\definecolor{yEdge}{HTML}{D4AC0D}   
\definecolor{yMid}{HTML}{F1C40F}    
\definecolor{yLight}{HTML}{F9E79F}  
\definecolor{yAccent}{HTML}{F4D03F} 
\definecolor{significantyellow}{HTML}{FFF8E1}

\newcommand{\bestavg}[1]{\cellcolor{lightyellow}\textbf{#1}}
\newcommand{\fastest}{\,\scriptsize\textcolor{orange}{\faBolt}}
\newcommand{\sigcell}{\cellcolor{lightyellow}}

  \lstdefinestyle{pythonstyle}{
    backgroundcolor=\color{backcolour},   
    commentstyle=\color{codegreen},
    keywordstyle=\color{codepurple},
    numberstyle=\tiny\color{codegray},
    stringstyle=\color{codepurple},
    basicstyle=\ttfamily\scriptsize,
    breakatwhitespace=false,         
    breaklines=true,                 
    captionpos=b,                    
    keepspaces=true,                 
    numbers=none,                    
    showspaces=false,                
    showstringspaces=false,
    showtabs=false,                  
    tabsize=2,
    frame=single
}
\title{LLM-Based Instance-Driven Heuristic Bias In the Context of a Biased Random Key Genetic Algorithm}

\author[1]{Camilo Chacón Sartori~\orcidlink{0000-0002-8543-9893}}
\author[2]{Martín Isla Pino}
\author[2]{Pedro Pinacho-Davidson~\orcidlink{0000-0001-9324-284X}}
\author[1]{Christian Blum~\orcidlink{0000-0002-1736-3559}}

\affil[1]{Artificial Intelligence Research Institute (IIIA-CSIC), Bellaterra, Spain.\protect\vspace{0.1cm} \texttt{\{cchacon, christian.blum\}@iiia.csic.es}}
\affil[2]{Department of Computer Science, University of Concepcion, Concepción, Chile.\protect\vspace{0.1cm} \texttt{\{misla2020, ppinacho\}@udec.cl}}

\begin{document}
\maketitle

\begin{abstract}
Integrating Large Language Models (LLMs) within metaheuristics opens a novel path for solving complex combinatorial optimization problems. While most existing approaches leverage LLMs for code generation to create or refine specific heuristics, they often overlook the structural properties of individual problem instances. In this work, we introduce a novel framework that integrates LLMs with a Biased Random-Key Genetic Algorithm (BRKGA) to solve the NP-hard Longest Run Subsequence problem. Our approach extends the instance-driven heuristic bias paradigm by introducing a human-LLM collaborative process to co-design and implement a set of computationally efficient metrics. The LLM analyzes these instance-specific metrics to generate a tailored heuristic bias, which steers the BRKGA toward promising areas of the search space. We conduct a comprehensive experimental evaluation, including rigorous statistical tests, convergence and behavioral analyses, and targeted ablation studies, comparing our method against a standard BRKGA baseline across 1,050 generated instances of varying complexity. Results show that our top-performing hybrid, BRKGA+Llama-4-Maverick, achieves statistically significant improvements over the baseline, particularly on the most complex instances. Our findings confirm that leveraging an LLM to produce an a priori, instance-driven heuristic bias is a valuable approach for enhancing metaheuristics in complex optimization domains.
\end{abstract}


\section{Introduction}\label{sec:intro}

The core challenge in combinatorial optimization is uncovering hidden patterns in complex, structured data. The emergence of Large Language Models (LLMs) such as GPT~\cite{openai2024gpt4technicalreport}, Llama~\cite{touvron2023llamaopenefficientfoundation}, and Gemini~\cite{geminiteam2024gemini15unlockingmultimodal} has transformed fields like natural language processing~\cite{minaee2025largelanguagemodelssurvey}, code generation~\cite{jiang2024surveylargelanguagemodels}, and reasoning over structured or tabular data~\cite{wu2025tabulardataunderstandingllms}. Beyond these tasks, LLMs can detect complex patterns and latent structures, perform symbolic reasoning, and extract task-relevant features that are often hard for humans to identify quickly~\cite{mirchandani2023largelanguagemodelsgeneral,wu2025tabulardataunderstandingllms}. It is this latent ability for abstract, data-driven reasoning on a large scale that remains largely unexplored for guiding sophisticated search algorithms. 

This pattern-recognition capability offers a significant opportunity to enhance metaheuristics (MHs)~\cite{10.1145/937503.937505}. Although MHs are well-suited for tackling NP-hard combinatorial problems, they typically operate without prior knowledge of the search space. While classic Machine Learning has been used to improve MHs~\cite{KARIMIMAMAGHAN2022393}, such integrations often require sophisticated model development and extensive training time. LLMs can mitigate these challenges by leveraging their pre-trained reasoning abilities to provide heuristic guidance in a zero-shot or few-shot context, without the need for traditional training~\cite{liu2024systematic}.

So far, the primary use of LLMs in the context of MHs has focused on code generation, allowing MHs to evolve their own code or be created from scratch~\cite{10752628,ye2024reevolargelanguagemodels}. In this work, we explore an alternative paradigm, originally developed in~\cite{10818476}: using the LLM as an \textit{instance-driven heuristic bias}. While a human expert can define relevant heuristic features, the LLM's role in this approach is to analyze the numerical matrix of these features for a specific instance---a pattern-recognition task often too complex for quick human assessment---to derive a quantitative, instance-specific search strategy. By converting this analysis into numerical parameters, the LLM guides a MH toward more promising regions of the search space. By doing so, our framework introduces a novel, a priori method for search guidance that complements established dynamic paradigms in evolutionary computation like hyper-heuristics. Our work strengthens this paradigm and expands it in several ways.

\subsection{Contribution and Paper Organization}\label{subsec:contribution}

The main contributions are:

\begin{itemize}
    \item We validate and extend the instance-driven paradigm by successfully applying it to the Longest Run Subsequence (LRS) problem---an NP-hard, string-based combinatorial optimization problem from a domain distinct from the framework's original context---demonstrating its potential for application to other combinatorial optimization problems.

    \item We introduce a co-design human–LLM framework for feature engineering. Instead of a purely expert-led process, our method uses the LLM to propose candidate metrics, while the human expert validates them for correctness and efficiency. This process yields instance-specific parameters that directly steer the search of a Biased Random-Key Genetic Algorithm (BRKGA).

    \item We present the first comprehensive validation of this framework on a problem distinct from its original social-network context. Our evaluation encompasses rigorous statistical analyses, a qualitative investigation of search behavior to elucidate how the guidance operates, and targeted ablation studies to systematically assess the contribution of each core component.

\end{itemize}

The paper unfolds as follows. Section~\ref{sec:background} reviews related work on LLMs in combinatorial optimization and metaheuristics for string-based problems, and introduces the LRS problem and the BRKGA. Our proposed framework is detailed in Section~\ref{sec:proposed_method}, followed by a comprehensive experimental analysis in Section~\ref{sec:experimental}. We then discuss our findings and their connection to the current literature in Section~\ref{sec:dis}, and conclude in Section~\ref{sec:conclusion}.

\section{Related Work and Background}
\label{sec:background}

The integration of MHs and LLMs is a new frontier in optimization. We begin this section with a discussion of existing efforts in this area before defining the specific problem we address. We then introduce our chosen metaheuristic, which provides the foundation for the next section: a detailed discussion of our novel integration framework.

\subsection{Metaheuristics and Large Language Models in Optimization Problems}
    \label{subsec:llm_optimization}

The application of LLMs to optimization began in 2023~\cite{yang2024largelanguagemodelsoptimizers,liu2024largelanguagemodelsevolutionary}, with their integration into metaheuristics emerging as a distinct research area in 2024. This period introduced several foundational paradigms, which we review here.

Integration strategies can be broadly divided into two axes: those enhancing the \textit{algorithm} as a static construct, and those adapting the search process to a specific \textit{problem instance}.

Three prominent paradigms target the algorithm itself:

\begin{itemize}
    \item \textbf{Discovery of Novel Heuristics:} This approach uses LLMs to discover mathematical functions or code snippets from scratch, aiming to produce a new, universally effective heuristic. A prominent example is FunSearch, which frames the problem as a search over a program space~\cite{Romera-Paredes2024, liu2024evolutionheuristicsefficientautomatic}.

    \item \textbf{Evolutionary Generation of Algorithms:} In this paradigm, an LLM acts as a sophisticated genetic operator within an evolutionary loop, treating algorithm prompts as individuals. The goal is to evolve a new, superior metaheuristic or component, as seen in LLaMEA~\cite{10752628} and ReEvo~\cite{ye2024reevolargelanguagemodels}. AlphaEvolve extends this to evolve entire algorithm populations~\cite{Novikov2025-vg}.
    
    \item \textbf{Direct Code Refinement:} Here, the LLM interacts directly with the source code of a functional metaheuristic to produce an improved version. It leverages its semantic understanding to identify inefficiencies or unused parameters, effectively creating a better, permanent version of the algorithm~\cite{sartori2025improvingexistingoptimizationalgorithms, sartori2025combinatorialoptimizationallusing}.
\end{itemize}

By contrast to these algorithm-centric approaches, a fourth paradigm redirects the focus:

\begin{itemize}
    \item \textbf{Instance-Driven Heuristic Generation:} Proposed in~\cite{10818476} and called OptiPattern, this approach employs the LLM not as a code generator or refiner, but as an analytical, pattern-recognition engine. The model analyzes precomputed metrics from a specific problem instance to produce \textit{ad hoc} heuristic guidance, uniquely tailored to that instance’s structure.
\end{itemize}

Our work builds upon the fourth, instance-driven paradigm, adapting and applying it for the first time, to our knowledge, to a string-based combinatorial optimization problem. We introduce a key enhancement to the original framework: while their approach relied on an expert to select instance metrics, we use the LLM to \textit{co-design the feature set} in a human-in-the-loop process. To implement this, we adapt the \textit{alpha-beta mechanism} proposed in the foundational paper, using it to translate the LLM's analysis of our co-designed metrics into a quantitative bias for the Biased Random-Key Genetic Algorithm (BRKGA) search, as detailed in Section~\ref{sec:proposed_method}.

\subsubsection{Positioning within Guided Search Paradigms in Evolutionary Computation (EC)}\label{sec:relation_ec}

Our framework, which uses an LLM as an instance-driven heuristic bias, can be understood by contrasting it with two established paradigms for guiding search in EC: hyper-heuristics~\cite{burke2013hyper} and parameter adaptation~\cite{10.1145/2996355}.\footnote{This point is noteworthy since the foundational work~\cite{10818476} overlooks it.}

\paragraph{Comparison with Hyper-heuristics.} Hyper-heuristics are high-level methodologies that manage a set of low-level heuristics. A common approach is a selection hyper-heuristic, which chooses the most appropriate low-level heuristic to apply at a given decision point from a predefined set. Our approach is different. The LLM does not select from a discrete set of known heuristics (e.g., ``choose heuristic A vs. heuristic B''). Instead, it generates a continuous, instance-specific parameterization for a single, flexible heuristic---the BRKGA's constructive decoder. In this sense, our work is closer to generative hyper-heuristics, but with a key distinction: the heuristic (the bias vector $\vec{L}$) is not generated through an evolutionary process or from grammatical building blocks, but is inferred in a single shot by a pre-trained model analyzing the problem instance's features.

\paragraph{Comparison with Parameter Adaptation.} Dynamic parameter adaptation methods adjust algorithm parameters during the run based on feedback from the search process (e.g., the success rate of an operator). Our LLM-guided approach, in contrast, is static and a priori. The analysis of the instance and the generation of the bias vector $\vec{L}$ occur entirely before the evolutionary search begins. This represents a trade-off: our method avoids the computational overhead of in-run adaptation and provides a strong initial ``compass'' for the search, but it cannot react to the search dynamics as they unfold.

Therefore, our framework occupies a unique niche. It acts as an a priori, instance-driven heuristic bias, leveraging the pattern-recognition capabilities of LLMs to create a bespoke search bias. This complements existing dynamic and selection-based approaches by offering a new way to inject deep, problem-specific knowledge into a metaheuristic before the first solution is even created.

\subsection{Metaheuristic Approaches for String-Based Optimization Problems}

String and substring combinatorial problems are fundamental in bioinformatics, largely because biological information---such as DNA, RNA, and proteins---is naturally represented as strings over a small alphabet~\cite{blum2016metaheuristics}. A significant portion of biological analysis involves searching for, comparing, or aligning subsequences, which can often be modeled as combinatorial optimization problems. Metaheuristics have proven particularly effective in this domain.

Some classic string-based problems where metaheuristics have been successfully applied include:

\begin{itemize}
    \item \textbf{Longest Common Subsequence (LCS) and its variants:} The objective is to find the longest subsequence common to a set of strings, where elements are not required to be contiguous~\cite{bergroth2000survey, 10.1145/322033.322044}. For the related Longest Common Square Subsequence (LCSqS) problem, Reixach et al.~\cite{reixach2024biased} used a BRKGA to find effective splitting points in the input string, reducing the problem to the standard LCS problem.

    \item \textbf{Shortest Common Superstring:} The goal is to find the shortest string that contains every string in a given set as a substring, a problem central to genome assembly~\cite{tarhio1988greedy}. To tackle this, Mousavi et al.~\cite{MOUSAVI2012457} developed a Beam Search algorithm incorporating a probabilistic mechanism.

    \item \textbf{Genome Rearrangement Problems:} This class of problems seeks the minimum number of large-scale operations (e.g., inversions, translocations) to transform one genome into another~\cite{fertin2009combinatorics}. Siqueira et al.~\cite{siqueira2021heuristics} demonstrated that techniques such as Local Search, Genetic Algorithms, and GRASP are highly effective for these tasks.
\end{itemize}

For a broader, recent analysis of the role of metaheuristics in bioinformatics, we refer the reader to the survey by Calvet et al.~\cite{calvet2023role}.

\paragraph{Integrating Metaheuristics and LLMs for Combinatorial Optimization in Bioinformatics.}
A recent systematic review on LLMs in combinatorial optimization~\cite[Table 7]{da2025large} shows that, within bioinformatics, applications have been almost exclusively limited to Enzyme Design—typically formulated on protein sequences but oriented toward structural and functional optimization. By contrast, most LLM integrations focus on domains such as Routing, Scheduling, and Network Design. Moreover, a recent survey on Evolutionary Computation and LLMs~\cite{chauhan2025evolutionarycomputationlargelanguage} does not mention any theoretical string problems being addressed. This gap highlights the novelty of our contribution: to the best of our knowledge, this is the first study to tackle a \textit{purely string-based} bioinformatics problem by integrating a metaheuristic with an LLM.

\subsection{The Longest Run Subsequence Problem}\label{subsec:lrs_problem}

Our work focuses on the Longest Run Subsequence (LRS) problem, a computationally challenging task first introduced by Dondi et al.~\cite{dondi_et_al:LIPIcs.CPM.2021.14} and more recently examined in~\cite{asahiro2023approximation,lai2024longest}. Given that the problem is NP-hard, metaheuristics like the BRKGA are a natural and effective fit~\cite{blum2025biasedrandomkeygenetic}. In this paper, however, we propose a new hybrid strategy to further enhance the performance of such algorithms.

A formal definition of the problem is as follows. Given an input instance $(S, \Sigma)$, where $S = s_1,s_2,\ldots,s_n$ is a string of length $n$ over a finite alphabet $\Sigma$, a string $S'$ is a subsequence of $S$ if $S'$ can be obtained by deleting zero or more characters from $S$. A subsequence $S'$ is a valid \textit{run-subsequence} if for any character $\sigma \in \Sigma$, all occurrences of $\sigma$ within $S'$ are contiguous blocks, also known as \textit{runs}. The objective of the LRS problem is to find a run-subsequence $S^*$ that has the maximum possible length.

\paragraph{Example.} Consider the input string $S = \text{ZZBCCZBBBC}$ over the alphabet $\Sigma = \{Z, B, C\}$. 
\begin{itemize}
    \item The subsequence $S' = \text{\underline{\textcolor{red}{Z}}B\underline{\textcolor{red}{Z}}C}$ is not valid because the characters `Z' are separated by `B'.
    \item The subsequence $S'' = \text{\underline{\textcolor{green!60!black}{Z}}\underline{\textcolor{green!60!black}{Z}}\underline{\textcolor{green!60!black}{B}}\underline{\textcolor{green!60!black}{C}}\underline{\textcolor{green!60!black}{C}}}$ qualifies as a valid run-subsequence, but it is not the longest possible.
\end{itemize}

\begin{figure}[t]
    \centering
    \includegraphics[width=1\linewidth]{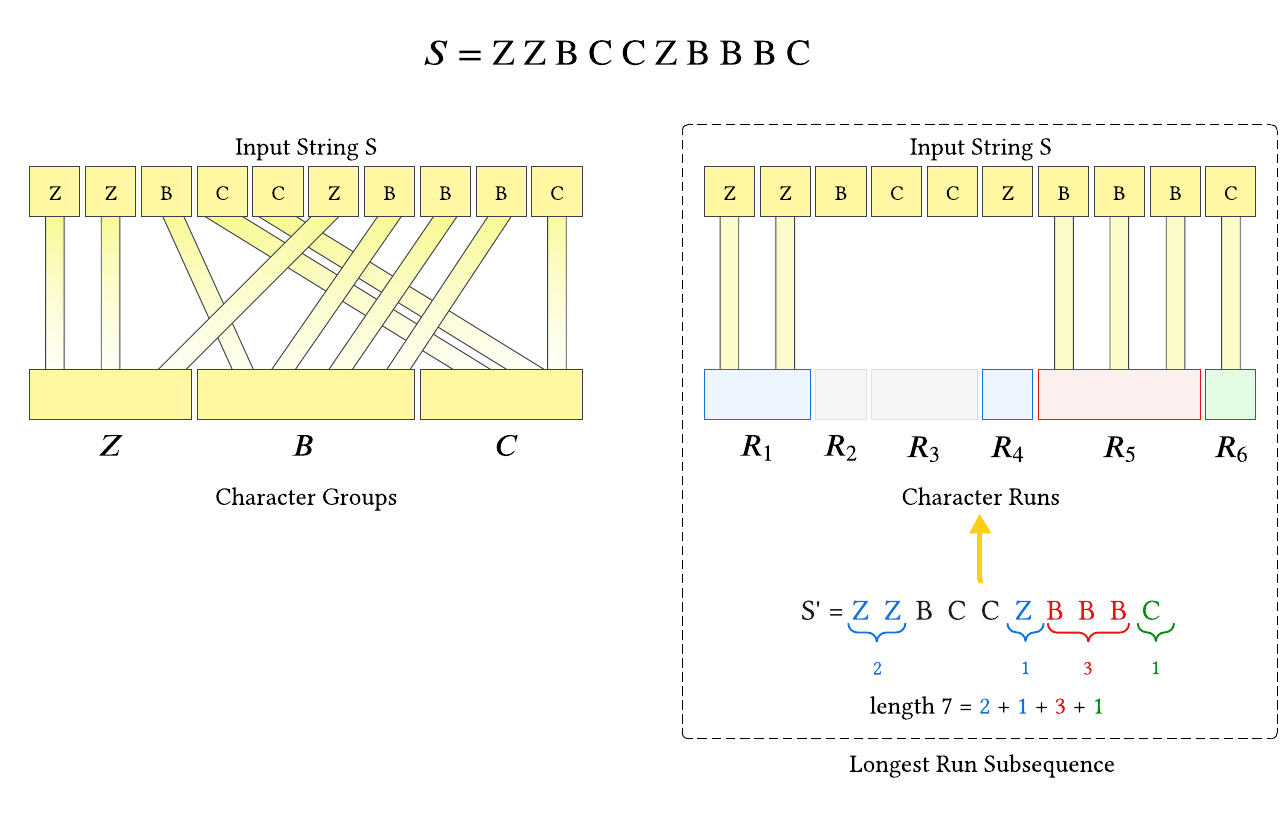}
  \caption{Example of the process to find a Longest Run Subsequence (LRS) in the string $S = \text{ZZBCCZBBBC}$. The left panel shows the grouping of characters into classes $Z$, $B$, and $C$, which serves as a precursor for identifying the LRS. The right panel depicts the reorganization of characters into runs ($R_1$, ..., $R_6$) and highlights the LRS $S' = \text{ZZZBBBC}$, with selected runs $R_1$, $R_4$, $R_5$, and $R_6$ contributing to a total run length of $2 + 1 + 3 + 1 = 7$.}

    \label{fig:LRS}
\end{figure}

As shown in Figure~\ref{fig:LRS}, the optimal solution for the string $S = \text{ZZBCCZBBBC}$ is the LRS $S^* = \text{ZZZBBBC}$, formed by selecting characters at positions 1, 2, 6, 7, 8, 9, and 10 (1-based indexing). This subsequence has length 7, and identical characters appear in contiguous blocks (ZZ, Z, BBB, C).

The LRS problem can be simplified by exploiting a structural property of the input string: each character $s_i$ belongs to a maximal substring $s[j,k] = s_j \ldots s_k$, where $1 \leq j \leq i \leq k \leq n$ and $s_j = \cdots = s_k = s_i$. These uniform substrings are called \emph{runs}. Notably, if an LRS solution includes any character from a run, it must include the entire run. Therefore, the input string can be compactly represented as a sequence of runs $R_1, R_2, \ldots, R_m$, where each run $R_i$ has character $c(R_i) \in \Sigma$ and length $l(R_i)$.

For $S = \text{ZZBCCZBBBC}$, the run decomposition is:
\begin{center}
\begin{tabular}{@{}cccc@{}}
\toprule
\textbf{Run} & \textbf{Character} & \textbf{Length} & \textbf{Original Positions} \\
\midrule
$R_1$ & Z & $l(R_1)=2$ & (positions 1--2) \\
$R_2$ & B & $l(R_2)=1$ & (position 3) \\
$R_3$ & C & $l(R_3)=2$ & (positions 4--5) \\
$R_4$ & Z & $l(R_4)=1$ & (position 6) \\
$R_5$ & B & $l(R_5)=3$ & (positions 7--9) \\
$R_6$ & C & $l(R_6)=1$ & (position 10) \\
\bottomrule
\end{tabular}
\end{center}

In general, let $\mathcal{R} = \{R_1, R_2, \ldots, R_m\}$ be the set of all runs. A valid solution corresponds to a subset $R' \subseteq \mathcal{R}$ such that, for any $R_i, R_k \in R'$ with $i < k$ and $c(R_i) = c(R_k)$, there is no $R_j \in R'$ with $i < j < k$ and $c(R_j) \neq c(R_i)$. In other words, runs with the same character must appear consecutively in the solution.

The objective is to find the subset $R^* \subseteq \mathcal{R}$ that maximizes the total run length:
\[
f(R^*) = \sum_{R_i \in R^*} l(R_i).
\]

In the case of the above example, $R^* = \{R_1, R_4, R_5, R_6\}$ yields the optimal solution $S^* = \text{ZZZBBBC}$ with $f(R^*) = 7$.

Henceforth, we represent an LRS instance as the triple $(S, \Sigma, \mathcal{R})$, where $\mathcal{R}$ is the run decomposition of $S$.

\FloatBarrier 
\subsection{Biased Random-Key Genetic Algorithm (BRKGA)}\label{subsec:brkga}

To tackle the LRS problem, we adopt the BRKGA. This choice is inspired by recent work~\cite{10818476}, where a hybrid of BRKGA and a LLM was successfully applied to a complex optimization task. The authors demonstrated that BRKGA’s random-key mechanism is particularly \textit{receptive to external guidance}: probabilities generated by the LLM can effectively steer the search process towards more promising regions of the solution space. Motivated by this potential, we adopt a similar hybrid strategy for the LRS problem.

A solution in BRKGA is encoded as a chromosome vector $\vec{v} \in [0,1]^m$, where $m$ is the total number of runs identified in the input string $S$. Each index $i$ in the vector corresponds directly to the $i$-th run, $R_i$. This problem-independent representation requires a problem-specific \textit{decoder}, which acts as a deterministic constructive heuristic. Its role is to translate the abstract random-key vector into a feasible LRS solution. It does so by first establishing a prioritized sequence of runs based on the keys, and then greedily constructing a solution by iterating through this sequence and adding only those runs that preserve subsequence validity. For the LRS problem, this means that selected runs must maintain their original relative order, and crucially, each run in the solution must consist of a unique character. The general pseudocode for this framework is shown in Algorithm~\ref{alg:brkga}.

Algorithm~\ref{alg:brkga} outlines this evolutionary process in detail. The procedure begins by generating an initial population $P$ of random-key vectors (line~2). Each chromosome is then decoded into a solution via the \Call{Decode}{} procedure and subsequently evaluated by the \Call{Evaluate}{} function (lines~2--3). The main evolutionary loop (lines~5--10) iteratively refines the population. In each generation, the next population is assembled from three distinct groups: top-performing individuals (\textit{elites}), randomly generated chromosomes (\textit{mutants}), and new \textit{offspring}. The offspring are generated via parameterized uniform crossover (line~7), where for each gene, the key is inherited from an elite parent with high probability $p_{elite}$, and from a non-elite parent otherwise. This mechanism strongly biases the search towards the characteristics of high-performing solutions while still incorporating diverse genetic material.

The inclusion of an \textit{elite} set (line~5) ensures monotonic progress, preventing degradation in the best-found solution across generations. Conversely, the use of \textit{mutants} (line~6) introduces crucial variation, injecting novel genetic material to avoid premature convergence to local optima. Newly generated individuals are then decoded and evaluated (lines~8--9), and the population is updated for the next iteration (line~10). This process continues until a termination criterion is met.

The core of our contribution lies in the modification of the \Call{Decode}{} function, where the LLM’s guidance is integrated into the search. If our hypothesis holds, this external information will help the decoder construct higher-quality solutions. Consequently, these solutions should receive higher scores from the \Call{Evaluate}{} function, effectively guiding the evolutionary process towards more promising areas of the solution space. This key component, highlighted in Algorithm~\ref{alg:brkga}, differentiates our hybrid method from the standard BRKGA and is fully detailed in Section~\ref{sec:proposed_method}.

\begin{algorithm}[!t]
\caption{The pseudocode of BRKGA for the LRS Problem}\label{alg:brkga}
\begin{algorithmic}[1] 
\Require a problem instance $(S, \Sigma, \mathcal{R})$, and values for parameters $p_{size}$, $p_e$, $p_m$, $prob_{elite}$

\State $P \gets \Call{GenerateInitialPopulation}{p_{size}, m}$ \Comment{$m$ is the number of runs in $S$}

\State \colorbox{lightyellow}{\Call{Decode}{P}} \Comment{Construct a solution from the keys of each chromosome}
\State \Call{Evaluate}{P} \Comment{Calculate the score of each constructed solution}

\While{computation time limit not reached}
    \State $P_e \gets$ \Call{EliteSolutions}{$P, p_e$}
    \State $P_m \gets$ \Call{Mutants}{$p_{size}, p_m, m$}
    \State $P_c \gets$ \Call{Crossover}{$P, p_e, prob_{elite}$}

    \State \colorbox{lightyellow}{\Call{Decode}{$P_m \cup P_c$}} \Comment{Construct solutions for new individuals}
    \State \Call{Evaluate}{$P_m \cup P_c$} \Comment{Calculate scores for new individuals}
    
    \State $P \gets P_e \cup P_m \cup P_c$
\EndWhile

\State \Return Best solution found in $P$
\end{algorithmic}
\end{algorithm}

\section{An LLM-Guided BRKGA for the LRS Problem}\label{sec:proposed_method}
\begin{figure}[t]
    \centering
    \includegraphics[width=\linewidth]{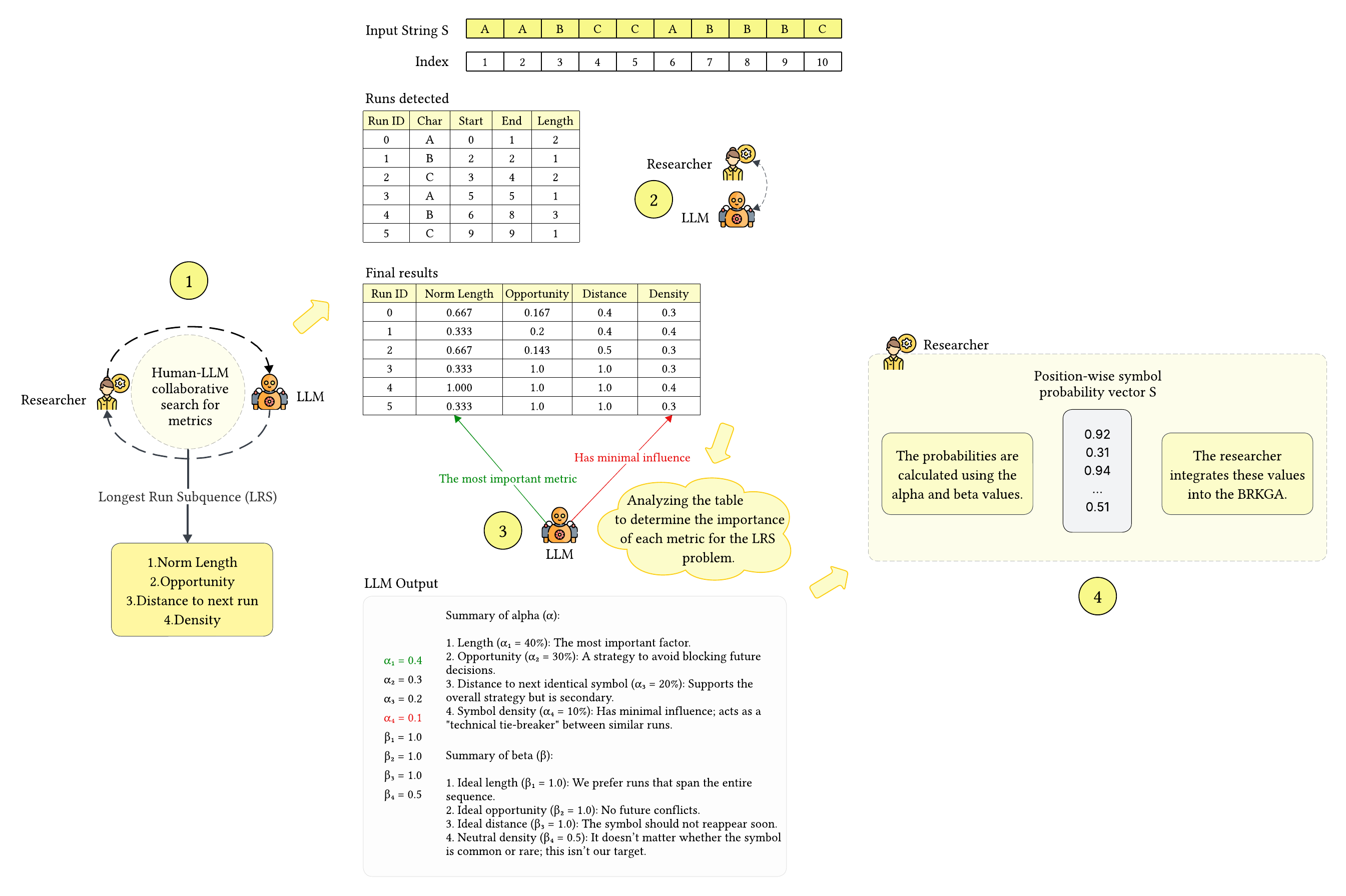}
    \caption{Overview of the proposed BRKGA-LLM framework for the LRS problem. The process consists of four phases: (1) A human-in-the-loop collaboration where an LLM suggests relevant LRS metrics and a researcher curates the final set. (2) The researcher develops a pipeline to extract these chosen metrics for each considered problem instance. (3) The LLM leverages its pattern-recognition capabilities to determine the relevance of each metric, assigning weights through the alpha-beta mechanism proposed in~\cite{10818476}. (4) Finally, these alpha-beta weights are used to generate selection probabilities for each run in the input string $S$, which in turn bias the BRKGA search process (see Algorithm~\ref{alg:brkga-llm-bias}).}
    \label{fig:overview}
\end{figure}

This section introduces the instance-driven heuristic bias framework, building on the foundational OptiPattern approach~\cite{10818476} and incorporating several methodological and performance-oriented enhancements. While OptiPattern demonstrated the potential of integrating a BRKGA with an LLM for network optimization problems, our work advances it in two key ways. First, we \textit{adapt} the instance-driven approach to string-based optimization while still leveraging a BRKGA. Second, we \textit{enhance} the methodology by introducing a novel collaborative process for discovering instance-specific metrics. The resulting framework consists of a four-phase process, illustrated in Figure~\ref{fig:overview}, which we briefly summarize as follows:
\begin{enumerate}
   \item \textbf{Metric Identification:} A collaborative phase where a human expert prompts the LLM to suggest candidate metrics for the problem, and then curates the final feature set.

   \item \textbf{Metric Extraction Process:} The development of a computational procedure that takes an input string, decomposes it into runs, and computes the value of each selected metric. For each metric, the corresponding extraction code is generated by an LLM and subsequently reviewed for correctness by a human expert.
   
   \item \textbf{LLM-Based Metric Prioritization:} The LLM analyzes the extracted data from an instance to assign an importance value to each metric, thereby quantifying its relevance.

   \item \textbf{LLM-Guided Integration into the BRKGA Decoder:} The generated values are used to compute a probability vector that biases the BRKGA's random keys, steering the heuristic search towards more promising solutions.
\end{enumerate}

We now describe each phase in detail, with an emphasis on reproducibility.

\subsection{Identifying Relevant Metrics}\label{subsec:phase_1}

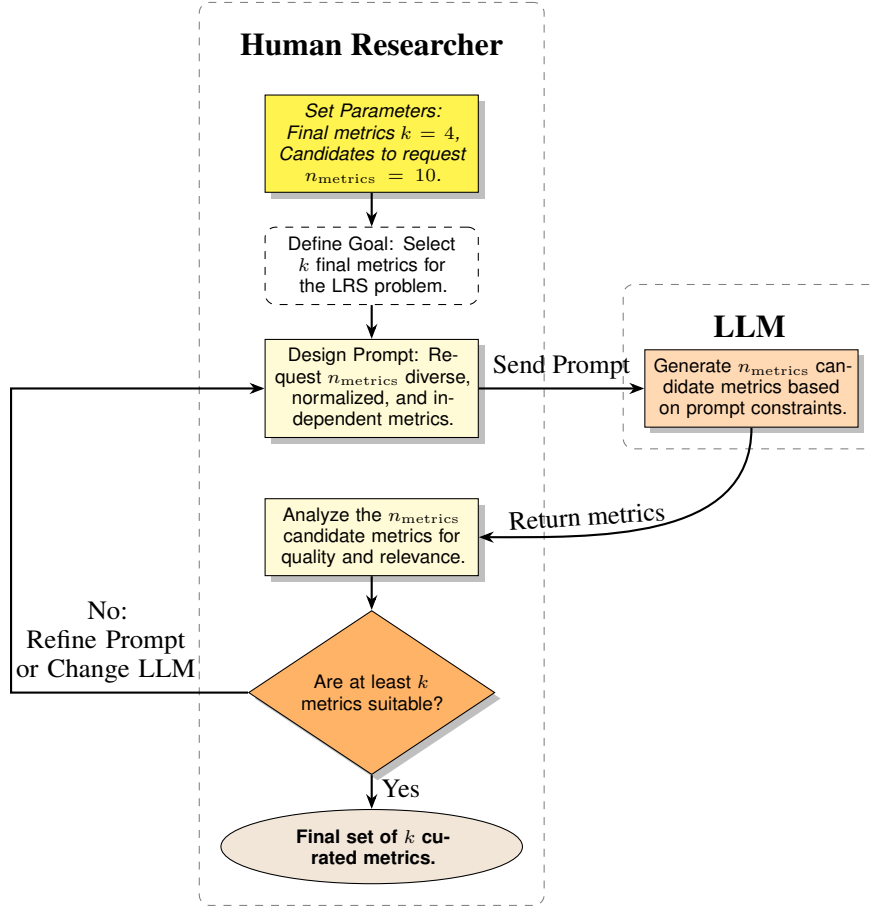
\begin{figure}[!ht]
    \centering
\begin{tikzpicture}[scale=0.75,
    node distance=0.45cm and 0.45cm,
    auto
]
\tikzstyle{actor} = [font=\large\bfseries]
\tikzstyle{parameter} = [
    rectangle, draw=black, fill=yellow!80, text width=2.6cm, text centered,
    minimum height=0.8cm, font=\sffamily\scriptsize\itshape, drop shadow
]
\tikzstyle{goal} = [
    rectangle, rounded corners, draw=black, dashed, text width=2.6cm,
    text centered, minimum height=0.8cm, font=\sffamily\scriptsize
]
\tikzstyle{human_task} = [
    rectangle, draw=black, fill=yellow!20, text width=2.6cm, text centered,
    minimum height=0.95cm, drop shadow, font=\sffamily\scriptsize
]
\tikzstyle{llm_task} = [
    rectangle, draw=black, fill=orange!30, text width=2.6cm, text centered,
    minimum height=0.95cm, drop shadow, font=\sffamily\scriptsize
]
\tikzstyle{decision} = [
    diamond, aspect=1.5, draw=black, fill=orange!60, text width=2.0cm,
    text centered, drop shadow, font=\sffamily\scriptsize
]
\tikzstyle{output} = [
    ellipse, draw=black, fill=brown!20, text width=2.6cm, text centered,
    minimum height=0.8cm, font=\sffamily\scriptsize\bfseries
]
\tikzstyle{flow_arrow} = [draw, -{Stealth[length=2mm]}, thick]

\node[actor] (human_actor) at (0,0) {Human Researcher};
\node[actor] (llm_actor) at (6.7,-5) {LLM};

\node[parameter, below=0.4cm of human_actor] (params) {Set Parameters: Final metrics $k=4$, Candidates to request $\nmetrics=10$.};
\node[goal, below=of params] (goal) {Define Goal: Select $k$ final metrics for the LRS problem.};
\node[human_task, below=of goal] (prompt) {Design Prompt: Request $\nmetrics$ diverse, normalized, and independent metrics.};
\node[human_task, below=0.8cm of prompt] (analyze) {Analyze the $\nmetrics$ candidate metrics for quality and relevance.};
\node[decision, below=of analyze] (decision_node) {Are at least $k$ metrics suitable?};
\node[output, below=of decision_node] (output) {Final set of $k$ curated metrics.};
\node[llm_task, right=2.2cm of prompt] (llm_gen) {Generate $\nmetrics$ candidate metrics based on prompt constraints.};

\node[draw=gray,dashed,inner sep=8pt,rounded corners,fit=(human_actor) (params) (goal) (prompt) (analyze) (decision_node) (output)] (lane_left) {};
\node[draw=gray,dashed,inner sep=8pt,rounded corners,fit=(llm_actor) (llm_gen)] (lane_right) {};

\path [flow_arrow] (params) -- (goal);
\path [flow_arrow] (goal) -- (prompt);
\path [flow_arrow] (prompt.east) -- node[above] {Send Prompt} (llm_gen.west);
\path [flow_arrow] (llm_gen.south) to[out=-90, in=0] node[above, sloped, pos=0.7] {Return metrics} (analyze.east);
\path [flow_arrow] (analyze) -- (decision_node);
\path [flow_arrow] (decision_node) -- node[right, pos=0.4] {Yes} (output);

\path [flow_arrow] (decision_node.west) -| 
    node[above, align=center, text width=2.8cm, pos=0.3] {No:\\Refine Prompt\\or Change LLM} ++(-4.2,0) |- (prompt);

\end{tikzpicture}

    \caption{Phase 1 of the BRKGA-LLM Framework: A human-in-the-loop process where the researcher first sets key parameters (target metrics $k$, candidates to request $\nmetrics$) and then interacts with an LLM to identify and select a suitable set.}
    \label{fig:phase_1}
\end{figure}

As shown in Figure~\ref{fig:phase_1}, this phase aims to identify a set of relevant metrics to guide the heuristic search. We employ a human-in-the-loop co-design process that begins by prompting an LLM to generate a diverse pool of $\nmetrics$ candidate metrics tailored to the LRS problem. Unlike~\cite{10818476}, which relied solely on expert-defined metrics, our approach uses LLMs to expand the search space beyond human-defined options.

The prompt---reproduced below---explicitly instructs the LLM to propose metrics that are normalized, independent, useful for guiding an algorithm, and, crucially, mutually distinct.

\begin{tcolorbox}[
    colback=yellow!3,      
    boxrule=0.6pt,       
    arc=1mm,
    sharp corners=south,
    enhanced,
    breakable,
    title=\textbf{Prompt for Metric Selection},
    fonttitle=\bfseries,
    fontupper=\ttfamily 
]
Propose \textbf{ten} independent, quantifiable metrics to guide the step-by-step construction of a solution for the Longest Run Subsequence (LRS) problem. The purpose of these metrics is to evaluate the heuristic value of selecting a single, candidate run ($R_i$) at a given stage of a constructive algorithm.

Each metric must capture a \emph{distinct and meaningful aspect} of the run's potential contribution or its associated risk, without overlapping conceptually or computationally with the others. The metrics should be:

\begin{enumerate}
    \item Fully independent from each other (no metric should subsume or depend on another).
    \item Expressed in a normalized or scale-invariant way where appropriate (e.g., relative to input size).
    \item Relevant to guiding a search algorithm, considering properties of the candidate run ($R_i$) in relation to the overall problem structure or a partially constructed solution.
\end{enumerate}

Avoid trivial variations of the same idea (e.g., raw vs. normalized length).
\end{tcolorbox}

Following the LLM’s output, the human researcher curates the list and selects the final $k \leq \nmetrics$ metrics for our framework. The human researcher curates this list based on two co-primary objectives: heuristic potential and computational efficiency. The expert's role is not just to identify metrics that seem informative, but to immediately discard those with high algorithmic complexity, ensuring the resulting feature set is viable for practical application. In our experiments, we set $\nmetrics = 10$ to maximize candidate diversity and $k = 4$ to maintain a compact set. As noted in~\cite{10818476}, the number of metrics has a direct impact on token count and, consequently, on API processing costs. For this reason, we selected a value close to the five metrics used in that study, a choice discussed in greater detail in Section~\ref{subsec:phase_3}, which describes the prompt construction for metric analysis.

As the flowchart shows in Figure~\ref{fig:phase_1}, if the $k$ metrics are deemed insufficient in quality, two refinement strategies are possible: (1) modifying the prompt, since LLMs are sensitive to query formulation; or (2) switching to a different LLM---while our experiments used GPT-4 (June 2025), any suitable model could be employed. This process forms a human-in-the-loop refinement cycle that continues until the researcher is satisfied, that is, until, after reviewing each metric’s definition, they can clearly determine its adequacy. This decision may be guided by several considerations, including the following ones:

\begin{itemize}
    \item \textbf{Computational efficiency:} This is the foremost criterion. Any metric that cannot be computed in near-linear time with respect to the input size is discarded, regardless of its perceived heuristic value. This ensures the framework remains practical for large-scale instances.
    \item \textbf{Distinctiveness:} Although the prompt specifies that metrics should be different, the researcher should ensure that the selected metrics are sufficiently distinct to enable the LLM, during the analysis stage, to capture diverse patterns. Choosing highly similar metrics would reduce this capacity and lead to unnecessary token expenditure.
\end{itemize}

It is important to note that our goal in this phase is not to identify the single optimal combination of metrics for the LRS problem. Exhaustively evaluating all combinations would be computationally prohibitive---not because of the prompt used in this phase, but due to the prompt generated later to analyze the metrics (see Section~\ref{subsec:phase_3}) and the number of BRKGA executions required to confidently state, \textit{``OK, this set of metrics is the best compared to all other metric combinations.''} A systematic search for optimal feature sets remains a promising avenue for future work, particularly as LLM token costs continue to decrease.

However, in the experimental part, we include an ablation study in Section~\ref{sec:valid_phase_1_2}, where we address the following questions: (1) why select $k = 4$ metrics instead of 3 or 5, and (2) what happens when $k$ metrics are randomly chosen from the set of $\nmetrics$ metrics.

\subsection{Metric Extraction Pipeline}\label{subsec:phase_2}

The second phase of our framework focuses on robust and accurate extraction of the selected metrics. This phase addresses a key challenge identified in recent literature: directly tasking LLMs with extracting features and performing calculations from raw optimization problem descriptions has been shown by~\citeauthor{da2024large}~\cite{da2024large} to be both computationally demanding and unreliable.

Following the approach advocated by~\cite{10818476} and later empirically validated by~\cite{da2024large}, we avoid using the LLM for direct feature computation. Instead, our framework maintains strict separation between data calculation and data analysis. All metric values for each instance are pre-computed through a human-validated computational pipeline, ensuring the correctness and reliability of the feature matrix $F$. This design shifts the LLM's role from low-level numerical computation to high-level pattern recognition, thereby improving both reliability and interpretability.

Having established this design rationale, we now provide the formal definition for each of the $k=4$ selected metrics. For a given input string $S$ of length $n= |S|$, we first decompose it into a set of runs $\mathcal{R} = \{R_1, \dots, R_m\}$. Each run $R_i$ is a tuple $(c_i, s_i, l_i)$, representing its character, start index, and length, respectively.

The formal definition for each of the $k=4$ selected metrics is now provided. It is noteworthy that all four of these curated metrics can be computed efficiently, with their algorithmic complexity being dominated by the initial run decomposition.

\begin{enumerate}
    \item \textbf{Normalized Length ($M_L$)}
    \begin{itemize}[leftmargin=*, topsep=2pt, itemsep=2pt]
        \item \textit{Rationale:} This is the most fundamental metric, as the objective of the LRS problem is to maximize total length. Normalizing by the maximum run length allows the model to assess the relative importance of a run within the context of the specific instance.
        \item \textit{Formal Definition:} Let $l_{max} = \max_{R_j \in \mathcal{R}} \{l_j\}$. The metric for a run $R_i$ is given by:
        $$ M_L(R_i) = \frac{l_i}{l_{max}} $$
    \end{itemize}

    \item \textbf{Opportunity ($M_O$)}
    \begin{itemize}[leftmargin=*, topsep=2pt, itemsep=2pt]
        \item \textit{Rationale:} A crucial strategic metric that looks ahead. Selecting a run might prevent the selection of many other runs later on. This metric quantifies how ``unconstraining'' a run is, favoring selections that keep future options open.
        \item \textit{Formal Definition:} Let $\Sigma_{>i} = \{c_j \mid s_j > (s_i + l_i - 1)\}$ be the set of distinct characters that appear in runs after $R_i$. The metric is defined as the fraction of the total alphabet $\Sigma$ that remains available:
        $$ M_O(R_i) = \frac{|\Sigma_{>i}|}{|\Sigma|} $$
    \end{itemize}

    \item \textbf{Distance to Next Run ($M_D$)}
    \begin{itemize}[leftmargin=*, topsep=2pt, itemsep=2pt]
        \item \textit{Rationale:} This metric helps identify fragmented ``super-runs.'' A short distance to the next occurrence of the same character indicates that two runs can be linked in a solution with minimal intermediate ``cost,'' making the current run more attractive.
        \item \textit{Formal Definition:} Let $e_i = s_i + l_i - 1$ be the end index of run $R_i$. Let $NextPos(c_i, e_i) = \min(\{k > e_i \mid S[k] = c_i\} \cup \{n\})$. The metric is the normalized distance to this next occurrence:
        $$ M_D(R_i) = \frac{NextPos(c_i, e_i) - e_i}{n} $$
    \end{itemize}

    \item \textbf{Global Character Frequency ($M_F$)}
    \begin{itemize}[leftmargin=*, topsep=2pt, itemsep=2pt]
        \item \textit{Rationale:} Formerly named ``Density,'' this metric provides global context about a run's character. It acts as a powerful tie-breaker, allowing the LLM to learn whether it is strategically better to prioritize runs of common or rare characters.
        \item \textit{Formal Definition:} Let $count(c_i) = |\{k \mid S[k] = c_i\}|$ be the total number of occurrences of character $c_i$ in $S$. The metric is:
        $$ M_F(R_i) = \frac{count(c_i)}{n} $$
    \end{itemize}
\end{enumerate}

\paragraph{Justification for discarded metrics.} Exemplary, we present some of the LLM-suggested metrics that were discarded, along with the reasons for their exclusion.

\begin{itemize}
    \item \textbf{External Fragmentation Potential:}
    \begin{itemize}[leftmargin=*, topsep=2pt, itemsep=2pt]
        \item \textit{Reason for Discarding:} Although it captures a different nuance of spacing, its strategic implication was deemed to overlap significantly with ``Opportunity'', without providing enough new information to justify its higher computational cost.
        \item \textit{Potential Formalism:} It could be defined as $M_{EFP}(R_i) = \frac{1}{n} \sum_{k=e_i+1}^{NextPos(c_i, e_i)-1} \mathbb{I}(S[k] \neq c_i)$, where $\mathbb{I}(\cdot)$ is the indicator function.
    \end{itemize}
    
    \item \textbf{Adjacency to Predecessor:}
    \begin{itemize}[leftmargin=*, topsep=2pt, itemsep=2pt]
        \item \textit{Reason for Discarding:} This metric is dynamic and solution-dependent, as it relies on the previously selected run. Our framework requires static, pre-computed metrics to generate a bias vector before the search begins.
        \item \textit{Potential Formalism:} For a candidate run $R_i$ and the previously selected run $R_p$, the metric would be $M_{AP}(R_i, R_p) = s_i - e_p$. Its dependence on $R_p$ makes it unsuitable.
    \end{itemize}

    \item \textbf{Internal Competitive Threat:}
    \begin{itemize}[leftmargin=*, topsep=2pt, itemsep=2pt]
        \item \textit{Reason for Discarding:} Its core concept was found to be a less effective variant of what is already captured by ``Normalized Length.''
        \item \textit{Potential Formalism:} It could be defined as $M_{ICT}(R_i) = l_i / \max\{l_j \mid c_j = c_i\}$, which normalizes a run's length only against other runs of the same character.
    \end{itemize}
\end{itemize}

The key difference between the selected and discarded metrics lies in their algorithmic efficiency. Metrics with high computational cost can drastically increase the preprocessing time in the context of large problem instances, as shown in the ablation study in Section~\ref{sec:valid_phase_1_2}. This consideration becomes even more critical given that LLMs often struggle to generate code that consistently satisfies strict time and space complexity requirements~\cite{chambon2025bigobenchllmsgenerate}.

\subsubsection{LLM-Assisted Code Generation for Metrics}

Once the set of chosen metrics is determined, we can collaborate again with the LLM to generate the implementation code. In particular, we make use of an LLM as a co-pilot for code generation. The objective is to implement the $k=4$ metrics selected in the previous phase. To facilitate this, we first manually implement the foundational \texttt{compute\_runs\_decomposition} function, shown in Listing~\ref{lst:runs_decomposition}, which takes an input string $S$ and returns a list of its constituent runs.

\begin{lstlisting}[language=Python, style=pythonstyle, % <-- Aplicar estilo aquí
    caption=Python function to decompose an input string into a list of runs.,
    label={lst:runs_decomposition}]
    
def compute_runs_decomposition(S):
    runs = []
    n = len(S)
    if n == 0:
        return runs
    current_letter = S[0]
    start = 0
    for i in range(1, n):
        if S[i] != current_letter:
            runs.append((current_letter, start, i - start, start, i-1))  # Added end position
            current_letter = S[i]
            start = i
    runs.append((current_letter, start, n - start, start, n-1))  # Added end position
    return runs
\end{lstlisting}

With this helper function providing the required \texttt{runs} data structure, we then prompt the LLM to generate the Python\footnote{Version3.10.} code for our four selected metrics: Normalized Length ($M_L$), Opportunity ($M_O$), Distance to Next Run ($M_D$), and Global Character Frequency ($M_F$). The code from Listing\ref{lst:runs_decomposition} is inserted into the dynamic placeholder \texttt{\{\{code\}\}} in the subsequent prompt.

\begin{tcolorbox}[
    colback=yellow!5,      
    boxrule=0.6pt,       
    arc=1mm,
    sharp corners=south,
    enhanced,
    breakable,
    title=\textbf{Prompt for Python Code Generation of Metrics},
    fonttitle=\bfseries,
    fontupper=\ttfamily 
]
You are an expert programmer specializing in heuristic algorithm design. Your task is to write four Python functions to calculate specific metrics for the Longest Run Subsequence (LRS) problem.

I will provide you with a Python function, \texttt{compute\_runs\_decomposition(S)}, that preprocesses an input string \texttt{S}. This function returns a list of tuples, where each tuple represents a 'run' and has the format \texttt{(character, start\_index, length, start\_pos, end\_pos)}.

Based on this, please implement the following four functions. Each function should take the list of \texttt{runs} and the original string \texttt{S} as input and return a list of numerical values, one for each run.

\begin{enumerate}
    \item \textbf{Normalized Length}: For each run, calculate its length divided by the maximum run length in the entire list.
    \item \textbf{Opportunity}: For each run, calculate the fraction of the total alphabet that appears in the string after the current run has ended.
    \item \textbf{Distance to Next Run}: For each run, find the distance from its end position to the start of the next run of the \textit{same character}. Normalize this distance by the total length of the string \texttt{S}. If the character does not appear again, the distance should be 1.0.
    \item \textbf{Global Character Frequency}: For each run, calculate the total number of times its character appears in the entire string \texttt{S}, normalized by the string's length.
\end{enumerate}

Ensure your code is efficient and well-documented. Do not repeat the \texttt{compute\_runs\_decomposition} function in your answer.

\textbf{\{\{code\}\}}
\end{tcolorbox}

With the Python functions for calculating each metric implemented, we proceed to construct the tabular data structure depicted in Phase 2 of Figure~\ref{fig:overview}. A script first processes the input string $S$ using the \texttt{compute\_runs\_decomposition} function to obtain the ordered list of runs, $\mathcal{R} = \{R_1, \dots, R_m\}$. Subsequently, each of the four metric functions ($M_L, M_O, M_D, M_F$) is applied to this data to generate the final table of feature values.

Formally, this process yields a feature matrix $F \in \mathcal{R}^{m \times 4}$, where $m$ is the number of runs. Each row $\vec{f_i}$ of this matrix is a feature vector corresponding to the run $R_i$ and is defined as:

\begin{equation}\label{eq:matrix_metrics}
\vec{f_i} = \langle M_L(R_i), M_O(R_i), M_D(R_i), M_F(R_i) \rangle
\end{equation}

This structured, quantitative representation of the problem instance is a critical prerequisite for the next phase, as it serves as the direct input for the primary prompt of our BRKGA-LLM framework.

\subsection{Prioritizing Metrics with the Alpha-Beta Mechanism}\label{subsec:phase_3}

After selecting the metrics and computing the tabular data $\vec{f_i}$ for a given LRS instance, we move on to Phase 3. At this stage, the core of the framework comes into play, leveraging LLMs to analyze numerical tabular data (metrics) following the alpha-beta approach described in~\cite{10818476}. We now detail the process for constructing the template prompt.

\subsubsection{Definition Template Prompt}

The prompt we have designed consists of three tags, defined as follows:
\begin{equation}\label{eq:prompt_analysis}
P := \text{prompt}(\text{Tag}_1,\text{Tag}_2,\text{Tag}_3) 
\end{equation}

where
\begin{itemize}
    \item  $\text{Tag}_1$ is the [PROBLEM] tag,
    \item  $\text{Tag}_2$ is the [EVALUATION SEQUENCE] tag, and
    \item  $\text{Tag}_3$ is the [RULES ANSWERING] tag.
\end{itemize}

This definition is consistent with~\cite[Eq.4]{10818476}, except that we omit the example-related tag ($\text{Tag}_2$ in the original equation), as preliminary experiments showed no improvement in the quality of the generated parameters. For the LRS, its inclusion simply provided no benefits, while its exclusion simplified the corresponding prompt $P$, whose template is shown in the following:

\begin{tcolorbox}[
    colback=yellow!5,      
    boxrule=0.6pt,       
    arc=1mm,
    sharp corners=south,
    enhanced,
    breakable,
    fonttitle=\bfseries,
    fontupper=\ttfamily 
]
[BEGIN PROBLEM]

The Longest Run Subsequence (LRS) problem is defined as follows: Given an input string S over an alphabet \(\Sigma\), the goal is to extract a subsequence $S^*$ composed of entire runs from S, such that each symbol from the alphabet appears in at most one run in $S^*$, and the total length of $S^*$ is maximized. A run is a maximal sequence of consecutive identical characters. The selected runs in $S^*$ must preserve their original order in S, and cannot overlap.

[END PROBLEM]

[BEGIN EVALUATION SEQUENCE]

Metrics description:

- Normalized\_length: Length of the run divided by the total string length.

- Opportunity: Estimated potential contribution of the run to the total LRS 1/(1+ gap), where gap= next\_run\_start - start.

- Distance\_next: Normalized distance to the next occurrence of the same symbol.

- Global Character Frequency: Frequency of the character in the entire string divided by its total length.

[BEGIN DATA]

node,normalized-length,opportunity ,distance-next,character-frequency 

\textbf{\{\{sequence\_data\_metrics\}\}}

[END DATA]

[END EVALUATION SEQUENCE]

[BEGIN RULES ANSWERING]

Consider the following equation to assign a probability range to each node: 
\begin{align*}
    \text{Influence} = \text{sigmoid}\big(
        &  alpha\_1 \cdot (1 - (beta\_1 - \text{normalized-length})) \\
        & + alpha\_2 \cdot (1 - (beta\_2 - \text{opportunity})) \\
        & + alpha\_3 \cdot (1 - (beta\_3 - \text{distance-next})) \\
        & + alpha\_4 \cdot (1 - (beta\_4 - \text{character-frequency}))
    \big)
\end{align*}
\begin{itemize}
    \item Alpha: Represents the weighting coefficients assigned to each metric. The sum of all alpha values must equal 1 (\(\sum_{i=1}^{4} \alpha_i = 1\)), and each alpha value (\(\alpha_{i}\)) is constrained to the range (0, 1). 
    \item Beta: Represents a factor of desirable results for each metric. Each beta value (\(\beta_{i}\)) is independent and constrained to the range (0, 1).
\end{itemize}

The response must be only in the following format:

alpha\_1=\textbf{\{\{value\_alpha\_1\}\}}

alpha\_2=\textbf{\{\{value\_alpha\_2\}\}}

alpha\_3=\textbf{\{\{value\_alpha\_3\}\}}

alpha\_4=\textbf{\{\{value\_alpha\_4\}\}}

beta\_1=\textbf{\{\{value\_beta\_1\}\}}

beta\_2=\textbf{\{\{value\_beta\_2\}\}}

beta\_3=\textbf{\{\{value\_beta\_3\}\}}

beta\_4=\textbf{\{\{value\_beta\_4\}\}}

[END RULES ANSWERING]
\end{tcolorbox}

The prompt is structured into three main sections, each demarcated by a specific tag, to guide the LLM's reasoning process:

\begin{itemize}
    \item \textbf{$\text{Tag}_1$: \texttt{[BEGIN PROBLEM]}} This section provides the LLM with the formal definition of the LRS problem, establishing the necessary context for the task.\footnote{The tendency would be to remove this tag as LLMs improve, since many LLMs---especially the more capable ones---are already familiar with the LRS problem. However, we use it considering more specialized LLMs that can run locally, which, being trained on less data, might lack deep knowledge about LRS.}

    \item \textbf{$\text{Tag}_2$: \texttt{[BEGIN EVALUATION SEQUENCE]}} This section first describes the four metrics and then provides the specific instance data within a nested \texttt{[BEGIN DATA]} tag. The data is presented as a feature vector $\vec{f_i}$ for each run (as defined in Eq.~\ref{eq:matrix_metrics}), 
    which is then inserted into the prompt via the dynamic placeholder \texttt{\{\{sequence\_data\_metrics\}\}}. To reduce the prompt’s token count, we follow~\cite[Section V-B]{10818476} and convert all numerical values to scientific notation, avoiding long floating-point representations.
    
    \item \textbf{$\text{Tag}_3$: \texttt{[BEGIN RULES ANSWERING]}} The final section presents the core task. It defines the influence equation that the LLM must parameterize by assigning appropriate \textit{alpha} (importance) and \textit{beta} (desirability) values.
\end{itemize}

The LLM's response to this prompt consists of eight specific values: four alpha ($\alpha$) importance weights and four beta ($\beta$) desirability factors, corresponding to the four metrics. The model is instructed to return this output in a simple key-value format, as specified at the end of the prompt (e.g., \texttt{alpha\_1=\{\{value\_alpha\_1\}\}}). The significance and function of these parameters are explained in Section~\ref{sec:alpha-beta}.

\subsubsection{LLM Execution Prompt}\label{sec:llm-execution}

We formalize the LLM’s task as the execution of a function, \texttt{execute}, which takes as input the LLM $LM$, a prompt $P$, and a set of inference hyperparameters $\Omega$. In this study, the hyperparameters were not tuned; values such as the temperature ($\Omega_{\text{temp}}$) were kept at their default settings. The execution is expressed as $\text{Output} := \text{execute}(LM, P, \Omega)$, producing a vector of four $\alpha$ and four $\beta$ coefficients, ${\alpha_1, \dots, \alpha_4, \beta_1, \dots, \beta_4}$, which are subsequently passed to the next phase of the framework.

We will now explain the role of the alpha-beta parameters.

\subsubsection{The Alpha-Beta Parameters}\label{sec:alpha-beta}

The central hypothesis of this work is that an LLM can discern meaningful patterns from subtle numerical variations within the feature matrix of a large-scale problem instance (e.g., an LRS sequence with `length=5000'). Based on this premise, we task the LLM with analyzing these features to determine a set of alpha and beta values that parameterize the influence model provided in the prompt, as depicted in Phase 3 of Figure~\ref{fig:overview}.

We now briefly define the alpha and beta parameters.\footnote{For a more exhaustive discussion, however, we refer the reader to the foundational work by Sartori et al.~\cite[Section IV-B]{10818476}.}

\paragraph{Alpha ($\alpha$) values.} A set of coefficients, $\alpha_i \in (0, 1)$, that represent the \textit{relative importance} or \textit{weight} the LLM assigns to each of the four heuristic metrics $F$. They are constrained to sum to one ($\sum_{i=1}^{4} \alpha_i = 1$), effectively forming a probability distribution that quantifies how much each metric should influence the final heuristic score. A higher $\alpha_i$ value signifies that the corresponding metric is considered more critical by the LLM in the decision-making process. In essence, they answer the question: ``How much does this metric matter?''

\paragraph{Beta ($\beta$) values.} A set of independent coefficients, $\beta_i \in (0, 1)$, that represent the \textit{target or ideal value} for each corresponding metric. Each beta serves as a benchmark against which a run's actual metric value is compared. For example, a $\beta$ value of 1.0 for the ``Opportunity'' metric indicates that the most desirable runs are those with no future conflicts. These values allow the LLM to define a ``perfect'' profile for a run, answering the question: ``What is the best possible value for this metric?''

We can therefore hypothesize that the length of the input string $S$ correlates with the amount of data available for the LLM's analysis. Longer sequences produce more runs, resulting in a larger feature matrix $F$ with more numerical values for the LLM to detect patterns from. Conversely, shorter sequences may not provide sufficient data for the model to discern meaningful trends. As we will show in the experimental section, our results confirm this hypothesis and further reveal that this pattern-detection capability is significantly more pronounced in larger-scale LLMs.

What makes this approach particularly noteworthy is that it is data-driven. The structure of each problem instance provides the necessary information to discover an ad-hoc heuristic, moving beyond the generalized solutions typically found through traditional trial-and-error or evolutionary methods.

\subsection{Integrating LLM Guidance into the BRKGA Decoder}\label{subsec:phase_4}

The final phase of our framework involves integrating the LLM-generated guidance into the BRKGA decoder, as depicted in Phase 4 of Figure~\ref{fig:overview}. First, the alpha and beta values obtained from the LLM are used to parameterize the influence equation from the prompt $P$. Following the methodology in~\cite{10818476}, a sigmoid function is applied to the output of this model; this transformation is known to improve convergence speed in genetic algorithms like BRKGA~\cite{4344687}. This process yields the final probability vector, $\vec{L}$, where each element $L_i$ represents the heuristic value or ``desirability'' of selecting the corresponding run $R_i$. This vector provides rich, instance-specific information to the BRKGA, enhancing its exploration of the search space.

With the bias vector $\vec{L}$ computed, we then modify the standard \Call{Decode}{} procedure from Algorithm~\ref{alg:brkga}. Our enhanced LLM-guided decoder is detailed in Algorithm~\ref{alg:brkga-llm-bias}. The key modification, highlighted in yellow, is the criterion used to order the runs for the greedy construction. Instead of relying solely on the random keys $\vec{v}$, our decoder biases the search by sorting according to the product $\vec{v}_i \cdot \vec{L}_i$. This mechanism prioritizes runs that the LLM has identified as heuristically promising, effectively steering the search towards unexplored and potentially higher-quality regions of the solution space.
\begin{algorithm}[ht]
\caption{Hybrid BRKGA-LLM Decoder for LRS with Emphasis on LLM Bias}
\label{alg:brkga-llm-bias}
\begin{algorithmic}[1]
\Procedure{Decode}{vector $\vec{v}$, vector $\vec{L}$}
    \State \textbf{Input:} Random-key vector $\vec{v} \in [0,1]^n$, LLM probability vector $\vec{L} \in [0,1]^m$.
    \State \textbf{Output:} A solution set of run indices $\mathcal{S}$ and its score.
    \State
    \Statex \textit{\small\textcolor{gray}{// Compute a permutation $\pi$ prioritizing runs favored by the LLM.}}
    \State \textbf{Note:} $\vec{L}$ introduces a \textbf{bias} that guides the greedy selection toward runs suggested by the LLM.
    \State Let $\pi$ be a permutation of $\{1, \dots, m\}$ such that for all $i \in \{1, \dots, m - 1\}$:
    \State \hspace{\algorithmicindent} \colorbox{lightyellow!40}{$\vec{v}_{\pi(i)} \cdot \vec{L}_{\pi(i)}$} $\ge$ \colorbox{lightyellow!40}{$\vec{v}_{\pi(i+1)} \cdot \vec{L}_{\pi(i+1)}$}
    \Statex \textit{\small\textcolor{gray}{// Here, the LLM bias $\vec{L}$ amplifies the preference of certain runs, effectively steering the search.}}
    \State
    \Statex \textit{\small\textcolor{gray}{// Greedily construct the solution set $\mathcal{S}$, following the LLM-biased ordering.}}
    \State $\mathcal{S} \gets \emptyset$
    \For{$i \gets 1$ to $m$}
        \Statex \textit{\small\textcolor{gray}{// Only add compatible runs; the LLM bias ensures promising candidates are considered earlier.}}
        \If{$\mathcal{S} \cup \{R_{\pi(i)}\}$ is a valid run-subsequence}
            \State $\mathcal{S} \gets \mathcal{S} \cup \{R_{\pi(i)}\}$
        \EndIf
    \EndFor
    \State
    \State score $\gets \sum_{j \in \mathcal{S}} \text{length}(R_j)$
    \State \Return $(\mathcal{S}, \text{score})$
\EndProcedure
\end{algorithmic}
\end{algorithm}

\subsection*{Example}

Returning to the example string in Figure~\ref{fig:LRS}, $S = \text{ZZBCCZBBBC}$, suppose that after Phase 3 the LLM produces the probability vector $\vec{L}$, where each element corresponds to one of the six runs corresponding to input string $S$:
\[
\vec{L} = \langle 0.85, 0.40, 0.70, 0.30, \textcolor{green!60!black}{\underline{\mathbf{0.95}}}, 0.25 \rangle
\]
The fifth element, corresponding to run $R_5$ (`BBB') in Figure~\ref{fig:LRS} (right panel), exhibits the highest probability, as the metrics indicate that this run is particularly relevant. This high value signals to the BRKGA that $R_5$ is promising, biasing the search toward it and accelerating convergence.

To see this mechanism in action, consider a random-key vector $\vec{v}$ generated for one individual in the population:
\[
\vec{v} = \langle 0.72, 0.91, 0.15, 0.88, 0.50, 0.65 \rangle
\]

\subsection*{Search Priority without LLM Guidance}
In a standard BRKGA, the decoder would sort the runs based solely on the random keys in $\vec{v}$. The run with the highest key is considered first.

\begin{itemize}
    \item \textbf{Priority based on $\vec{v}$}: The descending order of keys is $0.91 > 0.88 > 0.72 > 0.65 > 0.50 > 0.15$.
    \item \textbf{Run Order}: This corresponds to the run order (by ID) of $\langle R_2, R_4, R_1, R_6, \textbf{R}_5, R_3 \rangle$.
\end{itemize}

Notice that the heuristically promising run $R_5$ is ranked fifth. It would only be considered late in the greedy construction process, potentially after suboptimal choices have already been made.

\subsection*{Search Priority with LLM Guidance}
Our hybrid BRKGA-LLM decoder sorts the runs based on the biased keys, calculated as the element-wise product $\vec{v}_i \cdot \vec{L}_i$.

\begin{itemize}
    \item \textbf{Biased Key Calculation}:
    \begin{itemize}
        \item $R_1: 0.72 \cdot 0.85 = 0.612$
        \item $R_2: 0.91 \cdot 0.40 = 0.364$
        \item $R_3: 0.15 \cdot 0.70 = 0.105$
        \item $R_4: 0.88 \cdot 0.30 = 0.264$
        \item $\textbf{R}_5: 0.50 \cdot \textbf{0.95} = \textbf{0.475}$
        \item $R_6: 0.65 \cdot 0.25 = 0.163$
    \end{itemize}
    \item \textbf{Priority based on $\vec{v} \cdot \vec{L}$}: The new descending order is $0.612 > \textbf{0.475} > 0.364 > 0.264 > 0.163 > 0.105$.
    \item \textbf{New Run Order}: This corresponds to the run order of $\langle R_1, \textbf{R}_5, R_2, R_4, R_6, R_3 \rangle$.
\end{itemize}

Despite having a mediocre random key (0.50), the high probability assigned by the LLM (0.95) elevates the run $R_5$ from fifth to \textit{second place} in the construction priority. This demonstrates how the vector $\vec{L}$ effectively \textit{corrects} the random assignments of the keys, steering the search to evaluate heuristically valuable components much earlier. To validate that this LLM-generated guidance is not arbitrary, our experimental section will compare its performance against randomly generated bias vectors. Having detailed our framework, we now proceed to this analysis, examining how the vector $\vec{L}$ is influenced by instances of varying size and complexity.

\section{Experimental Analysis}\label{sec:experimental}

This section presents a comprehensive experimental evaluation of the proposed BRKGA-LLM framework. We first establish the methodological foundation by detailing the core components of our study: the portfolio of LLMs selected and their execution environment (Section~\ref{sec:llm_selection}); the benchmark instances used for testing (Section~\ref{sec:benchmark}); our general experimental setup (Section~\ref{sec:setup}); and the parameter tuning process used to ensure a fair comparison (Section~\ref{sec:tuning}). With this setup established, our analysis then explores the framework's performance across several key dimensions:

\begin{itemize}
    \item \textbf{Performance against Baseline:} We compare the solution quality and runtime of our hybrid approach against the standard BRKGA (Section~\ref{sec:perf_1}).

    \item \textbf{Validation of LLM Guidance:} We conduct a two-part ablation study. First, we validate the effectiveness of the co-designed metric set itself (Section~\ref{sec:valid_phase_1_2}). Second, we verify that the heuristic guidance generated by the LLM using these metrics is meaningful and not arbitrary (Section~\ref{sec:valid_phase_3}).

    \item \textbf{Behavioral Analysis:} We provide a visual analysis of the algorithm's search behavior to understand how the LLM influences exploration (Section~\ref{sec:analysis-behavior}).

    \item \textbf{Analysis of LLM-Generated Parameters:} We investigate the alpha-beta parameters learned by different LLMs to gain insight into their heuristic strategies (Section~\ref{sec:analysis-llm-parameters}).

    \item \textbf{Practical Considerations:} Finally, we analyze the API latency and associated economic costs of using different LLMs within our framework (Section~\ref{sec:analysis-api-latency}).
\end{itemize}

All results can be found in the repository: \url{https://github.com/martiniip/OptippaternLRS}.

\subsection{Selected Large Language Models and Execution Environment}\label{sec:llm_selection}

In contrast to the foundational work in~\cite{10818476}, our selection of LLMs was primarily driven by cost-effectiveness rather than solely by top performance, with a preference for open-weight models. As our framework's token consumption scales with instance size, API cost is a critical factor. Our portfolio of selected models, with rankings from the LMArena~\cite{chiang2024chatbotarenaopenplatform} Leaderboard\footnote{\url{https://lmarena.ai/}} as of June 2025, is as follows:
\begin{itemize}
    \item \textbf{GPT-4.1-mini} (ranked \#21) and \textbf{Gemini-2.5-Flash} (ranked \#10): Two high-performing, proprietary models. Public details on their architecture and parameter counts are not available.
    \item \textbf{Llama-3.2-3b} (ranked \#126): A small, open-weight model chosen specifically for its free-to-use status.
    \item \textbf{Llama-4-Maverick} (ranked \#51): A powerful, low-cost, open-weight model built on a Mixture-of-Experts (MoE) architecture with 128 experts, 17 billion active parameters per forward pass, and 400 billion total parameters.
\end{itemize}

All LLM interactions for Phase 3 of our method (Section~\ref{subsec:phase_3}) were managed via the OpenRouter API\footnote{\url{https://openrouter.ai/}}, which provides a unified interface for invoking diverse models. We utilized the default inference parameters provided by the platform for each model, ensuring no model-specific tuning was performed. The inclusion of the lower-ranked Llama-3.2-3b was a deliberate choice to benchmark the performance of a completely free-to-use model, presenting a valuable contrast to the foundational work where all selected models incurred costs. The cost associated with each LLM is further detailed in Section~\ref{sec:analysis-api-latency}.

\subsection{Benchmark Instances}\label{sec:benchmark}

As there is no publicly available set of benchmark instances in the recent literature on the LRS problem~\cite{dondi_et_al:LIPIcs.CPM.2021.14,asahiro2023approximation,lai2024longest}, we developed our own benchmark, emphasizing coverage across a wide range of controlled complexities. To this end, we generated thirty random instances for each combination of length $n \in \mathcal{L} = \{100, 200, 300, 500, 1000, 2000, 5000\}$ and alphabet size $|\Sigma| \in \mathcal{A} = \{2, 4, 8, 16, 32\}$, resulting in a total of $N=1050$ instances.

Notably, an increase in the alphabet size $|\Sigma|$ significantly raises the instance's complexity. While the number of runs tends to decrease, they become shorter and more fragmented, increasing the combinatorial ``noise'' and making it more challenging to find high-quality solutions. The algorithm used to generate these instances is detailed in Algorithm~\ref{alg:instance_generator}.

\begin{algorithm}[ht]
\caption{Instance Generation Algorithm}\label{alg:instance_generator}
\begin{algorithmic}[1]
\Require A set of target lengths $\mathcal{L}$, a set of alphabet sizes $\mathcal{A}$, and a number of repetitions $reps$.
\For{each alphabet size $|\Sigma| \in \mathcal{A}$}
    \State Define alphabet $\Sigma = \{c_1, \dots, c_{|\Sigma|}\}$
    \State Create a uniform probability distribution $P$ where $P(c_i) = 1/|\Sigma|$ for all $c_i \in \Sigma$.
    \For{each length $n \in \mathcal{L}$}
        \For{$k \gets 1$ to $reps$}
            \State Initialize an empty string $S$.
            \For{$j \gets 1$ to $l$}
                \State Randomly sample a character $c$ from $\Sigma$ according to $P$.
                \State Append $c$ to $S$.
            \EndFor
            \State Save $S$ to a new file named \texttt{len\_}$n$\texttt{\_sigma\_}$|\Sigma|$\texttt{\_}$k$\texttt{.txt}.
        \EndFor
    \EndFor
\EndFor
\end{algorithmic}
\end{algorithm}

\subsection{Experimental Setup}\label{sec:setup}
All experiments were conducted on a high-performance computing cluster consisting of 31 identical servers, each equipped with an Intel Xeon E3-1270 v6 (3.8 GHz) processor, 64 GB of RAM, and no dedicated GPUs. To guarantee a fair comparison, the baseline BRKGA and all BRKGA+LLM variants were executed on this same hardware.  This approach ensures that performance improvements can be attributed to the algorithmic changes rather than to differences in computational resources.

\subsection{Parameter Tuning}\label{sec:tuning}

To ensure a fair comparison, the parameters of all five algorithm variants---the baseline BRKGA and the four BRKGA+LLM hybrids---were independently tuned using the \texttt{irace} package~\cite{irace}, a well-known tool for automatic algorithm configuration. The tuning process was carried out with a total budget of 3000 executions for each algorithm. To this end, we created a dedicated tuning set consisting of 42 instances. Specifically, we generated one random instance for each combination of sequence length $n \in \{100, 200, 300, 500, 1000, 2000, 5000\}$ and alphabet size $|\Sigma| \in \{2, 4, 8, 16, 32\}$, yielding 35 instances in total. Additionally, we included one instance for each sequence length with a randomly chosen alphabet, resulting in 42 instances overall.

The parameter search space given to \texttt{irace} is detailed in Table~\ref{tab:irace_tuning}~(a). The final, optimized configurations selected by \texttt{irace} for each of the five algorithms are presented in Table~\ref{tab:irace_tuning}~(b). It is worth noting that with a budget of 3000 runs over 70 instances, a simple approach could test approximately 42 complete configurations. However, \texttt{irace} employs racing mechanisms to discard poorly performing configurations early, allowing for a much more efficient exploration of the search space within the same budget.

A clue we can infer from Table~\ref{tab:irace_tuning}~(b) is that the parameter selection in BRKGA+LLMs differs among the models, indicating that the probabilities generated by each LLM are not identical and affect the way BRKGA operates. We explore this in detail in the following sections.

\begin{table}[ht]
\centering
\caption{Parameter tuning summary for the BRKGA algorithm using irace.}
\label{tab:irace_tuning}

\begin{subtable}{0.9\textwidth}
    \centering
    \caption{Parameter space defined for the BRKGA algorithm.}
    \label{tab:irace_parameters_greek_desc}
    \begin{tabular}{lcc}
        \toprule
        \textbf{Parameter} & \textbf{Range} & \textbf{Type} \\
        \midrule
        Population Size ($\psi$) & (10, 400) & Integer \\
        Elite Proportion ($\pi_e$) & (0.1, 0.25) & Real \\
        Mutant Proportion ($\pi_m$) & (0.1, 0.3) & Real \\
        Elite Inheritance Probability ($\rho_e$) & (0.51, 0.8) & Real \\
        \bottomrule
    \end{tabular}
\end{subtable}

\vspace{1cm} 

\begin{subtable}{\textwidth}
    \centering
    \caption{Best configurations found by irace for the standard BRKGA and its LLM-assisted variants.}
    \label{tab:irace_configurations_greek_desc}
    \sisetup{table-format=1.2} 
    \begin{tabular}{
        c 
        l
        S[table-format=2.0]
        S[table-format=1.2]
        S[table-format=1.2]
        S[table-format=1.2]
    }
        \toprule
        & {\textbf{Algorithm Variant}} & {\textbf{$\psi$}} & {\textbf{$\pi_e$}} & {\textbf{$\pi_m$}} & {\textbf{$\rho_e$}} \\
        \midrule
        & Standard BRKGA & 16 & 0.25 & 0.17 & 0.56 \\
        \midrule 
        \multirow{4}{*}{\rotatebox{90}{BRKGA +}} 
        & GPT-4.1-mini & 36 & 0.23 & 0.21 & 0.68 \\
        & Gemini-2.5-Flash & 21 & 0.17 & 0.23 & 0.60 \\
        & Llama-3.2-3b & 27 & 0.19 & 0.21 & 0.78 \\
        & Llama-4-Maverick & 17 & 0.12 & 0.30 & 0.69 \\
        \bottomrule
    \end{tabular}
\end{subtable}

\end{table}

\subsection{Performance Comparison against Baseline}\label{sec:perf_1}

This section presents the empirical results of our BRKGA-LLM framework, demonstrating its effectiveness compared to a standard BRKGA baseline. The detailed results are reported in Table~\ref{tab:brkga_vs_llm}, which shows the performance for each combination of instance length ($L$) and alphabet size ($|\Sigma|$). In particular, each algorithm variant was applied exactly once to each problem instance. The numbers shown in the result tables are, therefore, averages over 30 problem instances of the same type. The execution time limit depends on the instance length, set to $n/5$ seconds. For example, instances with input string length $100$ are allocated $20$ seconds, whereas instances with input string length $5000$ are allocated $1000$ seconds. Cells highlighted in light yellow indicate the best average solution quality for that instance group, 
while the lightning bolt icon ({\scriptsize\textcolor{orange}{\faBolt}}) marks the fastest average execution time.

For simple instances ($n=100$ with $|\Sigma| \in \{2, 4\}$), performance is similar across algorithms. With larger alphabets, BRKGA+LLM variants consistently outperform the baseline across all lengths. However, as complexity increases with larger alphabet sizes, the BRKGA+LLM variants consistently deliver superior solutions across all lengths. This trend is especially evident for the BRKGA+Llama-4-Maverick integration, which achieves the best solution quality in nearly all cases; in the six rows where it does not rank first, its average is less than one point below the winner. Regarding computational time, it is noteworthy that BRKGA+LLM variants based on smaller LLMs such as Gemini-2.5-Flash and Llama-3.2-3b often find better solutions than the baseline in less average time.

The summary row at the bottom of Table~\ref{tab:brkga_vs_llm} confirms these findings in aggregate: on average, all four BRKGA+LLM variants outperform the baseline in both solution quality and computational efficiency. Notably, Gemini-2.5-Flash provides information that enables BRKGA to reach promising regions of the search space more quickly, even though its solution quality is lower than that of Llama-4-Maverick. Llama-4-Maverick, in particular, exhibits outstanding overall performance---an observation we will now validate through a robust statistical analysis.

\subsubsection{Statistical Justification}

We performed a two-part statistical analysis to rigorously validate our results.

\begin{figure}[ht]
    \centering
    \includegraphics[width=1\linewidth]{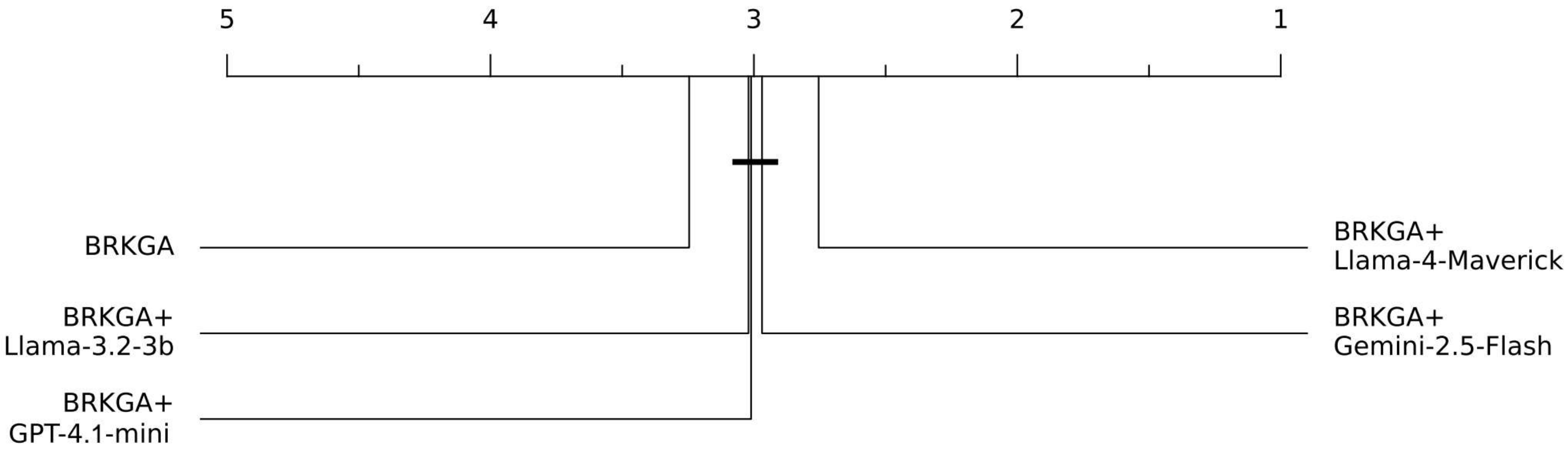}
\caption{Critical Difference (CD) diagram of the average ranks of five BRKGA variants evaluated on the complete dataset of $1050$ instances. The results highlight BRKGA+Llama-4-Maverick as the best-performing variant, whereas the BRKGA baseline consistently ranks below all hybrids. Horizontal bars indicate groups of algorithms whose performance differences are not statistically significant under the Nemenyi post-hoc test ($\gamma = 0.05$).}

    \label{fig:cd}
\end{figure}

\begin{enumerate}
    \item \textbf{Group Comparison:} A global comparison across all $N = 1050$ instances was first conducted. The Critical Difference diagram in Figure~\ref{fig:cd}, based on a Friedman test with a Nemenyi post-hoc analysis, reveals that BRKGA+Llama-4-Maverick attains the best (lowest) average rank, standing apart from all other variants with a statistically significant advantage. A second group---BRKGA+Gemini-2.5-Flash, BRKGA+GPT-4.1-mini, and BRKGA+Llama-3.2-3b---shows no statistically significant differences within it, but performs worse than BRKGA+Llama-4-Maverick. Finally, the baseline BRKGA ranks last, with significantly poorer results than any BRKGA+LLM configuration.
 
    To understand these results in more detail, a focused analysis for each group of instances is presented in Table~\ref{tab:statistical_summary_final}. This granular view confirms BRKGA+Llama-4-Maverick's robustness, as it consistently ranks in the top group for complex instances (i.e., $|\Sigma| \in \{16, 32\}$). Crucially, for the challenging $(n=300, |\Sigma|=32)$ group, BRKGA+Llama-4-Maverick is the only algorithm to achieve statistically superior performance. Collectively, these results reveal a clear trend: not only does BRKGA+Llama-4-Maverick's advantage become more pronounced as the alphabet size grows, but all LLM-hybrids consistently outperform the baseline on these more complex instances.

    \item \textbf{Specific Comparison (BRKGA+Llama-4-Maverick vs. Baseline BRKGA):} To precisely quantify the advantage of our best-performing algorithm variant, we conducted a direct pairwise comparison between BRKGA+Llama-4-Maverick and the standard BRKGA using the Wilcoxon signed-rank test (Table~\ref{tab:wilcoxon_significant}). The results confirm that our method with Llama-4-Maverick outperforms the baseline across the dataset, with its superiority becoming statistically significant in 42.86\% of the instance groups, primarily as the alphabet size $|\Sigma|$ increases. This finding aligns with the raw performance data, where the methods are indistinguishable for simple instances (e.g., $|\Sigma| \in \{2, 4\}$). This suggests that the LLM's guidance has its greatest impact when the problem’s combinatorial complexity is high and heuristic information becomes more critical.
    
    However, it is important to acknowledge that in 57.14\% of the instance groups, our framework's best representative did not achieve a statistically significant improvement over the baseline. This indicates that there still might be room for improvement, either by incorporating more advanced LLMs or by refining specific phases of the prompts utilized in our framework.
\end{enumerate}

We conclude that the LLM’s output is not random or hallucinatory, but reflects consistent pattern detection shaped by our prompt design (see Section~\ref{subsec:phase_3}).
Nevertheless, to rigorously verify that the observed performance gains stem from the LLM’s contributions---and not from other components of the system---the following section presents an ablation study, a widely used methodology for isolating and assessing the true impact of specific elements in LLM-based algorithms.

\begin{table}[t]
\centering
\caption{Performance of five BRKGA variants for the LRS problem, comparing different LLM integrations based on average solution quality and runtime in seconds. The best-performing solution values are highlighted in light yellow. A lightning bolt icon ({\scriptsize\textcolor{orange}{\faBolt}}) indicates the fastest execution time in each row.}
\label{tab:brkga_vs_llm}
\begin{tabular}{l l S[table-format=3.2] S[table-format=4.2]| S[table-format=3.2] S[table-format=4.2] S[table-format=3.2] S[table-format=4.2] S[table-format=3.2] S[table-format=4.2] S[table-format=4.2] S[table-format=4.2]}
\toprule
\textbf{Length} & \textbf{$\Sigma$} & \multicolumn{2}{c}{\textbf{BRKGA}} & \multicolumn{8}{c}{\textbf{BRKGA integrates with}} \\
\cmidrule(lr){3-4} \cmidrule(lr){5-12}
& & \multicolumn{2}{c}{} & \multicolumn{2}{c}{\textbf{GPT-4.1-mini}} & \multicolumn{2}{c}{\textbf{Gemini-2.5-Flash}} & \multicolumn{2}{c}{\textbf{Llama-3.2-3b}} & \multicolumn{2}{c}{\textbf{Llama-4-Maverick}} \\
\cmidrule(lr){5-6} \cmidrule(lr){7-8} \cmidrule(lr){9-10} \cmidrule(lr){11-12}
& & {avg.} & {time} & {avg.} & {time} & {avg.} & {time} & {avg.} & {time} & {avg.} & {time} \\
\midrule
\multirow{5}{*}{100} & 2 & \cellcolor{lightyellow}{\textbf{59.03}} & 0.00{\,\scriptsize\textcolor{orange}{\faBolt}} & \cellcolor{lightyellow}{\textbf{59.03}} & 0.00{\,\scriptsize\textcolor{orange}{\faBolt}} & \cellcolor{lightyellow}{\textbf{59.03}} & 0.00{\,\scriptsize\textcolor{orange}{\faBolt}} & \cellcolor{lightyellow}{\textbf{59.03}} & 0.00{\,\scriptsize\textcolor{orange}{\faBolt}} & \cellcolor{lightyellow}{\textbf{59.03}} & 0.00{\,\scriptsize\textcolor{orange}{\faBolt}} \\
& 4 & \cellcolor{lightyellow}{\textbf{41.07}} & 0.03 & \cellcolor{lightyellow}{\textbf{41.07}} & 0.01{\,\scriptsize\textcolor{orange}{\faBolt}} & \cellcolor{lightyellow}{\textbf{41.07}} & 0.01{\,\scriptsize\textcolor{orange}{\faBolt}} & \cellcolor{lightyellow}{\textbf{41.07}} & 0.01{\,\scriptsize\textcolor{orange}{\faBolt}} & \cellcolor{lightyellow}{\textbf{41.07}} & 0.02 \\
& 8 & 33.87 & 1.28 & \cellcolor{lightyellow}{\textbf{33.90}} & 1.69 & 33.87 & 1.02 & 33.87 & 0.68{\,\scriptsize\textcolor{orange}{\faBolt}} & \cellcolor{lightyellow}{\textbf{33.90}} & 0.92 \\
& 16 & 34.53 & 4.65 & 34.57 & 3.43 & 34.57 & 3.40 & 34.53 & 4.61 & \cellcolor{lightyellow}{\textbf{34.60}} & 3.04{\,\scriptsize\textcolor{orange}{\faBolt}} \\
& 32 & 39.93 & 9.91 & 40.40 & 8.67 & 40.93 & 10.97 & 40.83 & 7.94{\,\scriptsize\textcolor{orange}{\faBolt}} & \cellcolor{lightyellow}{\textbf{41.10}} & 8.29 \\
\midrule
\multirow{5}{*}{200} & 2 & \cellcolor{lightyellow}{\textbf{115.17}} & 0.00{\,\scriptsize\textcolor{orange}{\faBolt}} & \cellcolor{lightyellow}{\textbf{115.17}} & 0.00{\,\scriptsize\textcolor{orange}{\faBolt}} & \cellcolor{lightyellow}{\textbf{115.17}} & 0.00{\,\scriptsize\textcolor{orange}{\faBolt}} & \cellcolor{lightyellow}{\textbf{115.17}} & 0.00{\,\scriptsize\textcolor{orange}{\faBolt}} & \cellcolor{lightyellow}{\textbf{115.17}} & 0.00{\,\scriptsize\textcolor{orange}{\faBolt}} \\
& 4 & \cellcolor{lightyellow}{\textbf{72.77}} & 0.38 & \cellcolor{lightyellow}{\textbf{72.77}} & 0.09 & \cellcolor{lightyellow}{\textbf{72.77}} & 0.06{\,\scriptsize\textcolor{orange}{\faBolt}} & \cellcolor{lightyellow}{\textbf{72.77}} & 0.22 & \cellcolor{lightyellow}{\textbf{72.77}} & 0.17 \\
& 8 & \cellcolor{lightyellow}{\textbf{55.17}} & 4.27 & 55.10 & 2.71 & \cellcolor{lightyellow}{\textbf{55.17}} & 3.21 & 55.10 & 3.75 & 55.13 & 1.87{\,\scriptsize\textcolor{orange}{\faBolt}} \\
& 16 & 50.73 & 16.55 & 50.83 & 12.33 & 50.87 & 11.80 & 50.83 & 13.24 & \cellcolor{lightyellow}{\textbf{51.03}} & 11.30{\,\scriptsize\textcolor{orange}{\faBolt}} \\
& 32 & 54.70 & 25.10 & 55.37 & 22.66 & 54.50 & 21.15{\,\scriptsize\textcolor{orange}{\faBolt}} & 55.13 & 23.02 & \cellcolor{lightyellow}{\textbf{55.87}} & 21.21 \\
\midrule
\multirow{5}{*}{300} & 2 & \cellcolor{lightyellow}{\textbf{165.47}} & 0.01 & \cellcolor{lightyellow}{\textbf{165.47}} & 0.00{\,\scriptsize\textcolor{orange}{\faBolt}} & \cellcolor{lightyellow}{\textbf{165.47}} & 0.00{\,\scriptsize\textcolor{orange}{\faBolt}} & \cellcolor{lightyellow}{\textbf{165.47}} & 0.00{\,\scriptsize\textcolor{orange}{\faBolt}} & \cellcolor{lightyellow}{\textbf{165.47}} & 0.00{\,\scriptsize\textcolor{orange}{\faBolt}} \\
& 4 & \cellcolor{lightyellow}{\textbf{102.73}} & 2.88 & \cellcolor{lightyellow}{\textbf{102.73}} & 0.65 & \cellcolor{lightyellow}{\textbf{102.73}} & 0.78 & \cellcolor{lightyellow}{\textbf{102.73}} & 0.63{\,\scriptsize\textcolor{orange}{\faBolt}} & \cellcolor{lightyellow}{\textbf{102.73}} & 0.74 \\
& 8 & 74.97 & 9.65 & 75.07 & 2.14{\,\scriptsize\textcolor{orange}{\faBolt}} & \cellcolor{lightyellow}{\textbf{75.10}} & 3.20 & 75.07 & 6.44 & \cellcolor{lightyellow}{\textbf{75.10}} & 4.17 \\
& 16 & 65.80 & 28.89 & 66.30 & 17.80{\,\scriptsize\textcolor{orange}{\faBolt}} & \cellcolor{lightyellow}{\textbf{66.77}} & 22.42 & 66.27 & 19.53 & 66.63 & 21.37 \\
& 32 & 65.70 & 40.60 & 66.03 & 33.47 & 66.50 & 40.22 & 66.47 & 35.35 & \cellcolor{lightyellow}{\textbf{67.63}} & 33.17{\,\scriptsize\textcolor{orange}{\faBolt}} \\
\midrule
\multirow{5}{*}{500} & 2 & \cellcolor{lightyellow}{\textbf{271.67}} & 0.02 & \cellcolor{lightyellow}{\textbf{271.67}} & 0.01 & \cellcolor{lightyellow}{\textbf{271.67}} & 0.00{\,\scriptsize\textcolor{orange}{\faBolt}} & \cellcolor{lightyellow}{\textbf{271.67}} & 0.01 & \cellcolor{lightyellow}{\textbf{271.67}} & 0.02 \\
& 4 & \cellcolor{lightyellow}{\textbf{162.77}} & 3.84 & 162.73 & 0.47 & \cellcolor{lightyellow}{\textbf{162.77}} & 0.84 & \cellcolor{lightyellow}{\textbf{162.77}} & 0.33{\,\scriptsize\textcolor{orange}{\faBolt}} & \cellcolor{lightyellow}{\textbf{162.77}} & 0.45 \\
& 8 & 109.57 & 11.51{\,\scriptsize\textcolor{orange}{\faBolt}} & 109.93 & 15.07 & \cellcolor{lightyellow}{\textbf{110.20}} & 14.01 & 110.13 & 13.29 & \cellcolor{lightyellow}{\textbf{110.20}} & 21.53 \\
& 16 & 90.57 & 49.01 & 90.60 & 30.33{\,\scriptsize\textcolor{orange}{\faBolt}} & 90.80 & 38.49 & 90.33 & 48.46 & \cellcolor{lightyellow}{\textbf{91.17}} & 42.28 \\
& 32 & 84.87 & 71.69 & 85.50 & 63.40 & 85.77 & 61.78{\,\scriptsize\textcolor{orange}{\faBolt}} & 86.07 & 63.51 & \cellcolor{lightyellow}{\textbf{87.17}} & 63.28 \\
\midrule
\multirow{5}{*}{1000} & 2 & \cellcolor{lightyellow}{\textbf{530.23}} & 0.05 & \cellcolor{lightyellow}{\textbf{530.23}} & 0.01{\,\scriptsize\textcolor{orange}{\faBolt}} & \cellcolor{lightyellow}{\textbf{530.23}} & 0.02 & \cellcolor{lightyellow}{\textbf{530.23}} & 0.05 & \cellcolor{lightyellow}{\textbf{530.23}} & 0.02 \\
& 4 & \cellcolor{lightyellow}{\textbf{301.10}} & 7.60 & \cellcolor{lightyellow}{\textbf{301.10}} & 6.91 & \cellcolor{lightyellow}{\textbf{301.10}} & 6.12{\,\scriptsize\textcolor{orange}{\faBolt}} & 301.00 & 8.75 & 301.00 & 6.18 \\
& 8 & 191.00 & 50.82 & \cellcolor{lightyellow}{\textbf{191.87}} & 47.87 & 191.23 & 33.07 & 191.23 & 32.18{\,\scriptsize\textcolor{orange}{\faBolt}} & 191.80 & 34.84 \\
& 16 & 142.70 & 119.45 & 142.90 & 99.31 & 143.33 & 99.68 & 142.77 & 95.64{\,\scriptsize\textcolor{orange}{\faBolt}} & \cellcolor{lightyellow}{\textbf{143.37}} & 107.15 \\
& 32 & 122.63 & 143.78 & 123.97 & 145.17 & 124.00 & 141.95 & 123.97 & 158.01 & \cellcolor{lightyellow}{\textbf{126.23}} & 140.86{\,\scriptsize\textcolor{orange}{\faBolt}} \\
\midrule
\multirow{5}{*}{2000} & 2 & \cellcolor{lightyellow}{\textbf{1041.73}} & 0.58 & \cellcolor{lightyellow}{\textbf{1041.73}} & 0.07{\,\scriptsize\textcolor{orange}{\faBolt}} & \cellcolor{lightyellow}{\textbf{1041.73}} & 0.12 & \cellcolor{lightyellow}{\textbf{1041.73}} & 0.15 & \cellcolor{lightyellow}{\textbf{1041.73}} & 0.12 \\
& 4 & 572.30 & 21.06 & 572.67 & 15.62 & \cellcolor{lightyellow}{\textbf{572.50}} & 6.18{\,\scriptsize\textcolor{orange}{\faBolt}} & \cellcolor{lightyellow}{\textbf{572.50}} & 22.29 & \cellcolor{lightyellow}{\textbf{572.50}} & 29.98 \\
& 8 & 343.27 & 115.35 & 343.17 & 84.56 & 343.10 & 104.27 & 343.70 & 117.35 & \cellcolor{lightyellow}{\textbf{343.90}} & 65.82{\,\scriptsize\textcolor{orange}{\faBolt}} \\
& 16 & 232.33 & 276.39 & 234.63 & 233.89 & 235.00 & 222.05 & 233.73 & 193.54{\,\scriptsize\textcolor{orange}{\faBolt}} & \cellcolor{lightyellow}{\textbf{235.10}} & 199.04 \\
& 32 & 182.90 & 303.72 & 186.80 & 300.61 & 188.13 & 320.46 & 186.23 & 299.06{\,\scriptsize\textcolor{orange}{\faBolt}} & \cellcolor{lightyellow}{\textbf{189.90}} & 317.78 \\
\midrule
\multirow{5}{*}{5000} & 2 & \cellcolor{lightyellow}{\textbf{2567.47}} & 0.90 & \cellcolor{lightyellow}{\textbf{2567.47}} & 1.00 & \cellcolor{lightyellow}{\textbf{2567.47}} & 0.50 & \cellcolor{lightyellow}{\textbf{2567.47}} & 0.27 & \cellcolor{lightyellow}{\textbf{2567.47}} & 0.22{\,\scriptsize\textcolor{orange}{\faBolt}} \\
& 4 & 1361.23 & 110.76 & \cellcolor{lightyellow}{\textbf{1361.30}} & 151.09 & 1361.17 & 92.46 & \cellcolor{lightyellow}{\textbf{1361.30}} & 117.40 & 1361.17 & 78.86{\,\scriptsize\textcolor{orange}{\faBolt}} \\
& 8 & 765.20 & 339.43 & 766.87 & 330.09 & \cellcolor{lightyellow}{\textbf{768.50}} & 313.15{\,\scriptsize\textcolor{orange}{\faBolt}} & 767.13 & 319.40 & 768.07 & 372.71 \\
& 16 & 478.63 & 801.50 & 483.57 & 630.08 & 481.33 & 571.54{\,\scriptsize\textcolor{orange}{\faBolt}} & 483.90 & 622.15 & \cellcolor{lightyellow}{\textbf{486.23}} & 742.38 \\
& 32 & 331.10 & 863.47 & 335.73 & 797.70 & 336.93 & 814.40 & 336.50 & 792.55{\,\scriptsize\textcolor{orange}{\faBolt}} & \cellcolor{lightyellow}{\textbf{338.40}} & 809.65 \\
\midrule
\multicolumn{2}{l}{\textbf{Average}} & 311.85 & 98.15 & 312.52 & 87.40 & 312.61 & 84.55{\,\scriptsize\textcolor{orange}{\faBolt}} & 312.53 & 86.28 & \cellcolor{lightyellow}{\textbf{313.07}} & 89.70 \\
\bottomrule
\end{tabular}
\end{table}

\begin{table}
    \centering
    \caption{Summary of the statistical analysis for the LRS problem, comparing algorithm performance across instance groups. 
    Rows are highlighted where the Friedman test detects a statistically significant difference ($p < 0.05$), and the final column lists the top-performing algorithms according to the post-hoc Nemenyi test.}
    \label{tab:statistical_summary_final}

    \sisetup{
        table-format=2.4,
        table-space-text-post={***},
        scientific-notation=true
    }

    \begin{tabular}{
        c
        c
        S
        S[table-format=1.4e-1, table-auto-round]
        c
        p{7cm}
    }
        \toprule
        \textbf{Length} & \textbf{$\Sigma$} & {\textbf{Statistic}} & {\textbf{P-Value}} & {\textbf{Signif.}} & \textbf{Best Algorithm(s) (Ranked)} \\
        \midrule
        \multirow{5}{*}{100} 
        & 2  & {---} & {---}   & No & -- \\
        & 4  & {---} & {---}   & No & -- \\
        & 8  & 4.0000 & 0.4060 & No & -- \\
        & 16 & 2.8000 & 0.5918 & No & -- \\
        & 32 & 10.5193 & \sigcell 0.0325 & \sigcell \textbf{Yes} & All algorithms are in the top group. \\
        \addlinespace
        \multirow{5}{*}{200} 
        & 2  & {---} & {---}   & No & -- \\
        & 4  & {---} & {---}   & No & -- \\
        & 8  & 2.2143 & 0.6964 & No & -- \\
        & 16 & 4.3465 & 0.3611 & No & -- \\
        & 32 & 14.2628 & \sigcell 0.0065 & \sigcell \textbf{Yes} & \textbf{1st:} Maverick, \textbf{2nd:} GPT-4.1-mini, \textbf{3rd:} Llama-3.2-3b \\
        \addlinespace
        \multirow{5}{*}{300} 
        & 2  & {---} & {---}   & No & -- \\
        & 4  & {---} & {---}   & No & -- \\
        & 8  & 4.6349 & 0.3268 & No & -- \\
        & 16 & 21.5625 & \sigcell 2.45e-4 & \sigcell \textbf{Yes} & \textbf{1st:} Gemini, \textbf{2nd:} Maverick, \textbf{3rd:} GPT-4.1-mini, \textbf{4th:} Llama-3.2-3b \\
        & 32 & 15.7383 & \sigcell 0.0034 & \sigcell \textbf{Yes} & \textbf{1st: Maverick} (statistically superior) \\
        \addlinespace
        \multirow{5}{*}{500} 
        & 2  & {---} & {---}   & No & -- \\
        & 4  & 4.0000 & 0.4060 & No & -- \\
        & 8  & 18.8792 & \sigcell 8.30e-4 & \sigcell \textbf{Yes} & All algorithms are in the top group. \\
        & 16 & 5.0374 & 0.2835 & No & -- \\
        & 32 & 12.4085 & \sigcell 0.0146 & \sigcell \textbf{Yes} & \textbf{1st:} Maverick, \textbf{2nd:} Llama-3.2-3b, \textbf{3rd:} Gemini, \textbf{4th:} GPT-4.1-mini \\
        \addlinespace
        \multirow{5}{*}{1000} 
        & 2  & {---} & {---}   & No & -- \\
        & 4  & 4.0000 & 0.4060 & No & -- \\
        & 8  & 9.5238 & \sigcell 0.0493 & \sigcell \textbf{Yes} & All algorithms are in the top group. \\
        & 16 & 3.5612 & 0.4686 & No & -- \\
        & 32 & 6.7379 & 0.1504 & No & -- \\
        \addlinespace
        \multirow{5}{*}{2000} 
        & 2  & {---} & {---}   & No & -- \\
        & 4  & 5.8667 & 0.2093 & No & -- \\
        & 8  & 4.1511 & 0.3859 & No & -- \\
        & 16 & 13.0410 & \sigcell 0.0111 & \sigcell \textbf{Yes} & All LLM variants are in the top group. \\
        & 32 & 12.1902 & \sigcell 0.0160 & \sigcell \textbf{Yes} & All LLM variants are in the top group. \\
        \addlinespace
        \multirow{5}{*}{5000} 
        & 2  & {---} & {---}   & No & -- \\
        & 4  & 3.4444 & 0.4864 & No & -- \\
        & 8  & 12.7574 & \sigcell 0.0125 & \sigcell \textbf{Yes} & All algorithms are in the top group. \\
        & 16 & 15.6103 & \sigcell 0.0036 & \sigcell \textbf{Yes} & All LLM variants are in the top group. \\
        & 32 & 7.7691 & 0.1004 & No & -- \\
        \bottomrule
    \end{tabular}

    \vspace{0.4cm}
    \begin{minipage}{0.6\textwidth}
        \centering
        \begin{tabular}{lc}
            \toprule
            \textbf{Summary Statistic} & \textbf{Value} \\
            \midrule
            Total groups analyzed & 35 \\
            Groups with significant differences & 11 \\
            Significance percentage & 31.4\% \\
            \bottomrule
        \end{tabular}

        \vspace{0.3cm}
        \footnotesize
        \begin{itemize}[leftmargin=*, itemsep=1pt]
            \item Algorithm names are shortened in the `Conclusion` column (e.g., `Maverick` for `BRKGA+Llama-4-Maverick`).
            \item Significance level (alpha) is set at 0.05.
            \item \colorbox{lightyellow}{Yellow highlighting} indicates a statistically significant Friedman test result for that group.
        \end{itemize}
    \end{minipage}
\end{table}

\begin{table}[ht]
    \centering
    \caption{Instance groups where \textbf{BRKGA+Llama-4-Maverick} demonstrated a statistically significant superiority over the baseline \textbf{BRKGA}. Significance is determined by a one-sided Wilcoxon signed-rank test ($p < 0.05$).}
    \label{tab:wilcoxon_significant}
    \sisetup{
        table-format=3.4,
        table-space-text-post={***}
    }
    
    \begin{tabular}{
        c 
        c 
        S
        S[table-format=1.4] 
    }
        \toprule
        \textbf{Length} & \textbf{$\Sigma$} & {\textbf{Statistic}} & {\textbf{P-Value}} \\
        \midrule
        100  & 32 & 22.0000 & 0.0014 \\
        \addlinespace
        200  & 16 & 26.0000 & 0.0363 \\
             & 32 & 62.5000 & 0.0019 \\
        \addlinespace
        300  & 16 & 18.5000 & 0.0068 \\
             & 32 & 73.0000 & 0.0014 \\
        \addlinespace
        500  & 8  & 11.5000 & 0.0073 \\
             & 32 & 83.0000 & 0.0053 \\
        \addlinespace
        1000 & 8  & 17.0000 & 0.0118 \\
             & 32 & 99.5000 & 0.0091 \\
        \addlinespace
        2000 & 8  & 45.0000 & 0.0358 \\
             & 16 & 52.5000 & 0.0003 \\
             & 32 & 63.5000 & 0.0003 \\
        \addlinespace
        5000 & 8  & 52.5000 & 0.0015 \\
             & 16 & 64.5000 & 0.0005 \\
             & 32 & 84.0000 & 0.0019 \\
        \bottomrule
    \end{tabular}
    
    \vspace{0.5cm}
    
    \begin{tabular}{lc}
        \toprule
        \textbf{Summary Statistic} & \textbf{Value} \\
        \midrule
        Total Groups Analyzed & 35 \\
        Groups with Significant Llama-4-Maverick Superiority & 15 \\
        Superiority Percentage & 42.86\% \\
        \bottomrule
    \end{tabular}

\end{table}

\subsubsection{Convergence Analysis}

While the previous sections examined average solution quality and total runtime, a crucial question remains regarding convergence behavior: Does the heuristic guidance from our framework influence the speed of convergence? In other words, do the LLM-hybrids not only achieve better solutions than the baseline, but also reach high-quality solutions faster?

\begin{figure}[t]
\centering

\begin{subfigure}{0.49\textwidth}
    \centering
    \includegraphics[width=\linewidth]{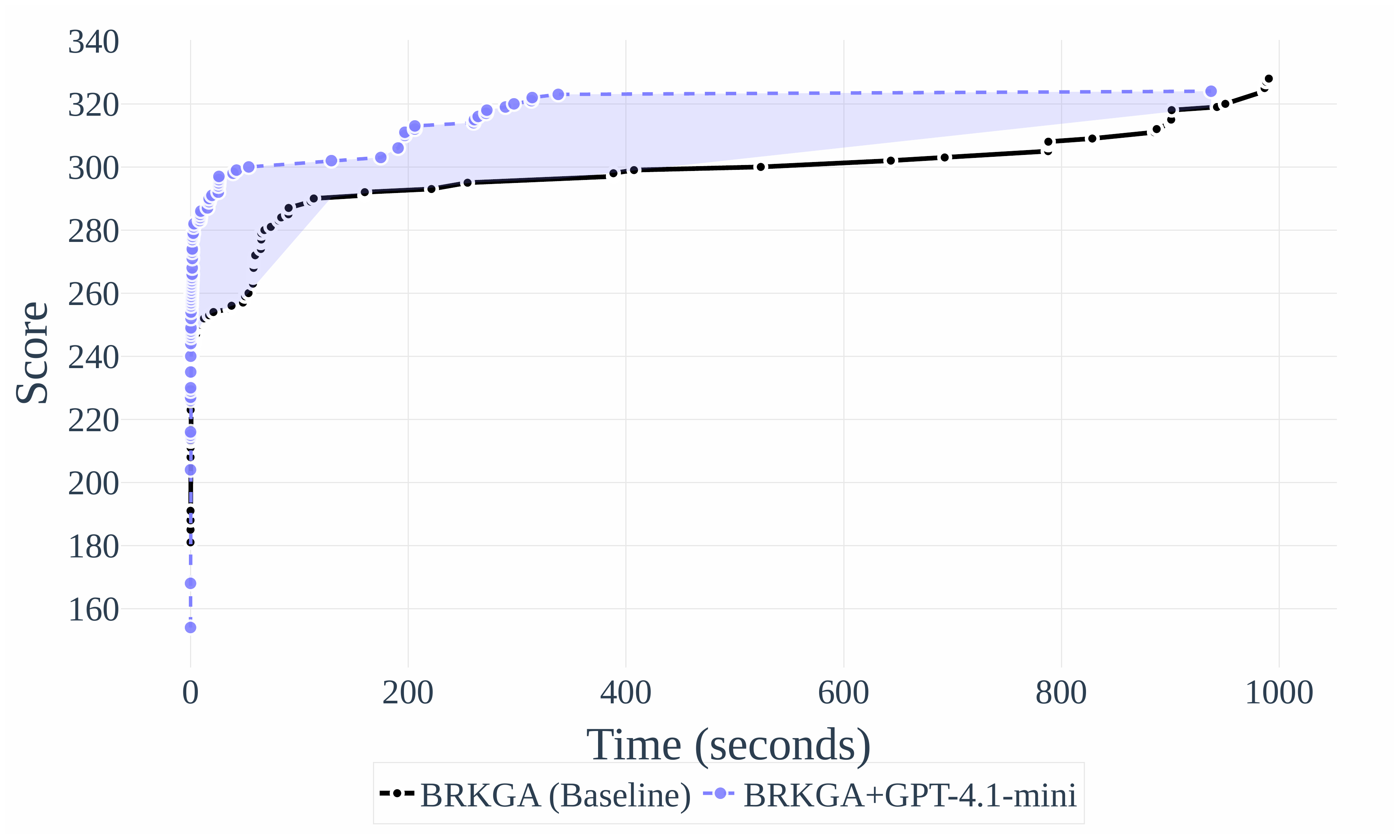}
    \caption{Analysis regarding BRKGA+GPT-4.1-mini.}
    \label{fig:sub_analysis_gpt}
\end{subfigure}
\hfill 
\begin{subfigure}{0.49\textwidth}
    \centering
    \includegraphics[width=\linewidth]{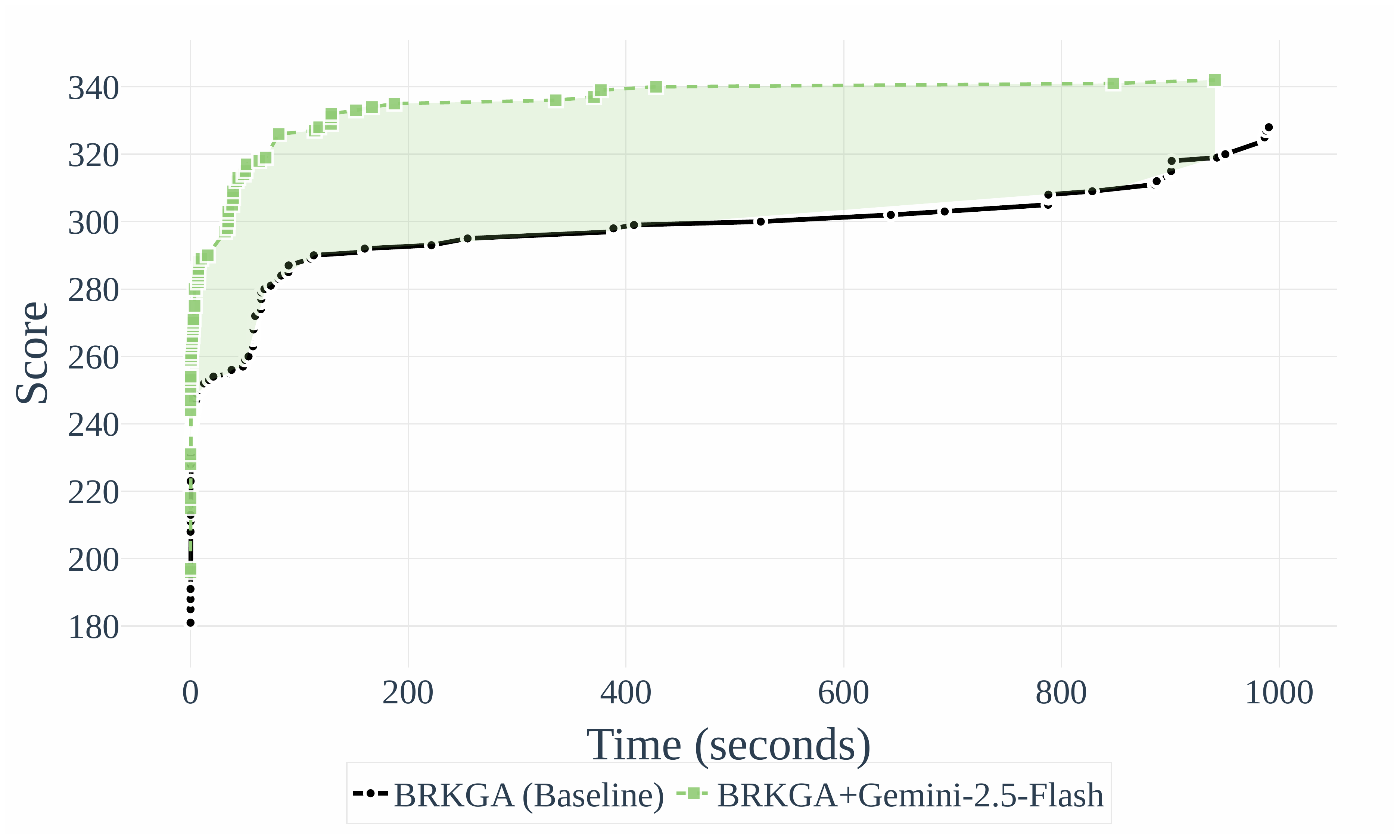}
    \caption{Analysis regarding BRKGA+Gemini-2.5-Flash.}
    \label{fig:sub_analysis_gemini}
\end{subfigure}

\vspace{0.5cm} 

\begin{subfigure}{0.49\textwidth}
    \centering
    \includegraphics[width=\linewidth]{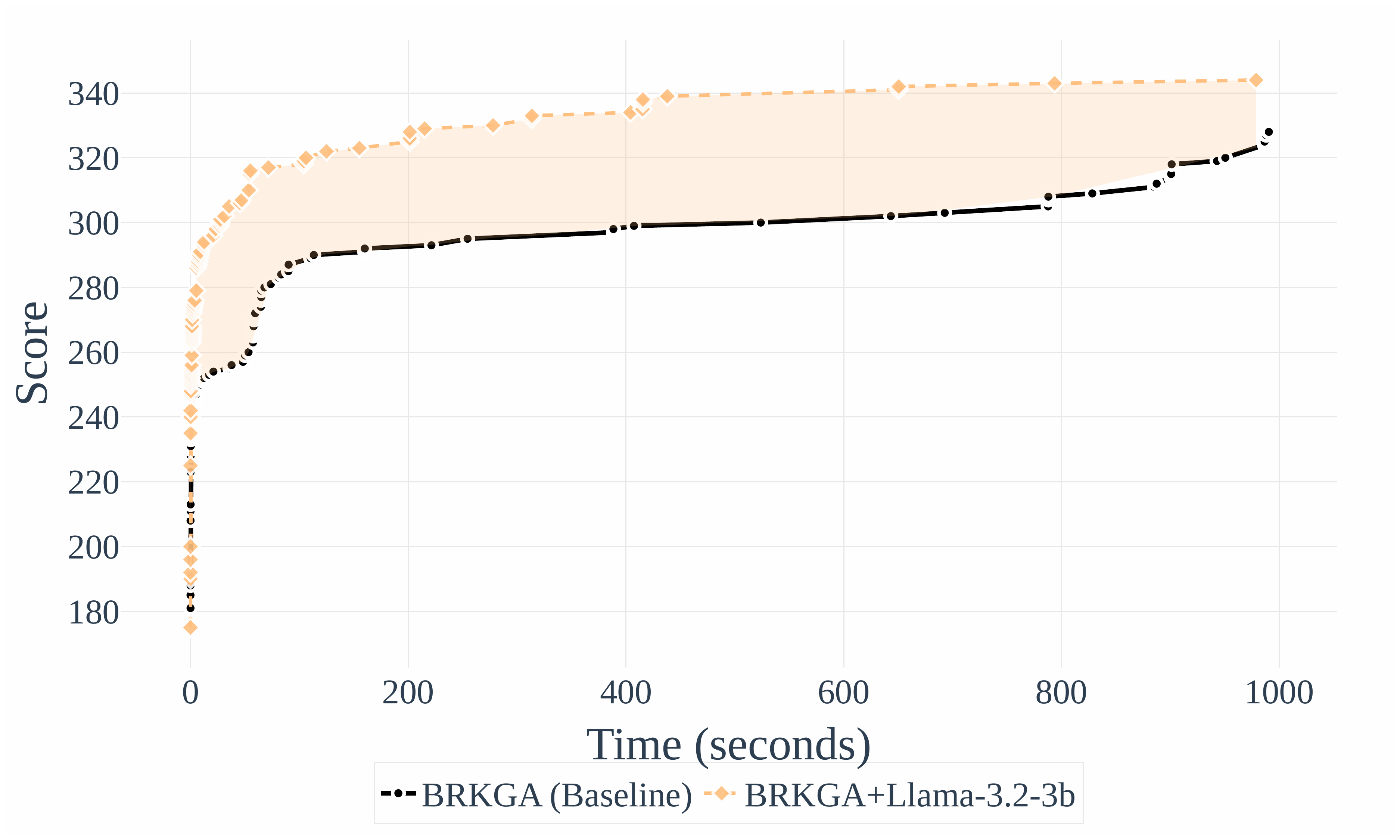}
    \caption{Analysis regarding BRKGA+Llama-3.2-3b.}
    \label{fig:sub_analysis_scout}
\end{subfigure}
\hfill 
\begin{subfigure}{0.49\textwidth}
    \centering
    \includegraphics[width=\linewidth]{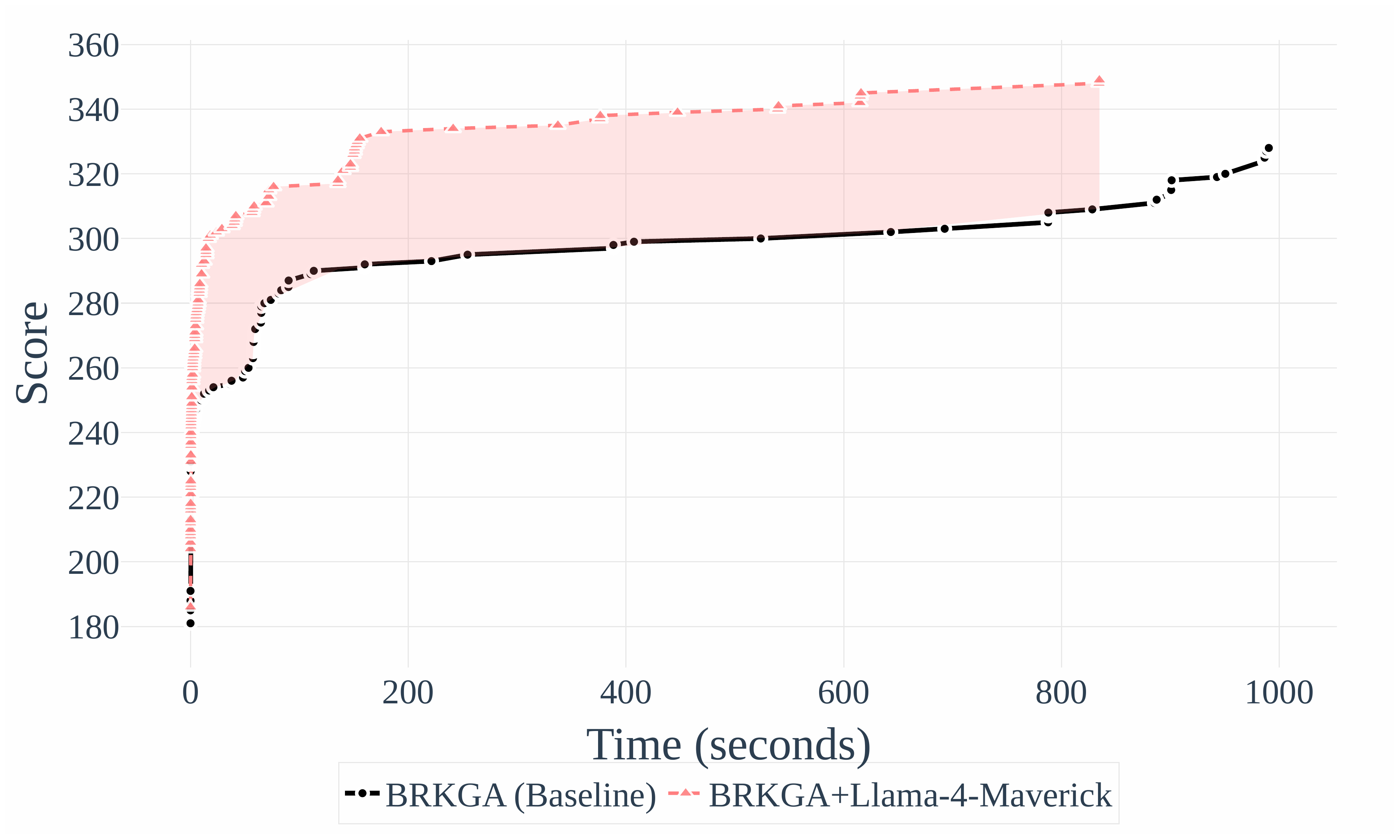}
    \caption{Analysis regarding BRKGA+Llama-4-Maverick.}
    \label{fig:sub_analysis_maverick}
\end{subfigure}

\caption{Convergence speed comparison of the four LLM-hybrids on a challenging ($n=5000$, $|\Sigma|=32$)-instance.}

\label{fig:llm_instance_analysis}

\end{figure}

Figure~\ref{fig:llm_instance_analysis} illustrates this convergence behavior for the four BRKGA+LLM variants on a challenging ($n=5000$, $|\Sigma|=32$)-instance with a CPU time limit of 1000 seconds. The performance of the standard BRKGA is included in each of the four subplots as a common baseline for comparison. Consistent with the final results from Table~\ref{tab:brkga_vs_llm}, the plots clearly show that all LLM-guided variants converge to better solutions more rapidly than the baseline.

A more detailed analysis, however, reveals notable differences in the search dynamics among the LLM variants. Both BRKGA+GPT-4.1-mini and BRKGA+Gemini-2.5-Flash (subplots (a) and (b)) exhibit strong initial improvements but tend to stagnate in suboptimal regions after roughly 400 seconds. In contrast, both Llama-based models (subplots (c) and (d)) continue to find improved solutions for much longer, with BRKGA+Llama-4-Maverick showing a particularly steep upward trend in solution quality. Interestingly, the convergence curve for GPT-4.1-mini is different from the other ones, showing poor performance even compared to the baseline in this specific case. It is critical to note, however, that this is an outlier; the aggregate results confirm that GPT-4.1-mini's overall performance is superior to the baseline. Ultimately, these individual trajectories highlight that while all hybrids converge faster than the standard BRKGA, their search dynamics and susceptibility to premature convergence vary significantly.

\subsection{Validation of the Co-Designed Metric Set}\label{sec:valid_phase_1_2}

Our framework's primary distinction from the approach in~\cite{10818476} is the use of an LLM to propose an initial set of $m$ candidate metrics, which are then curated by a human expert. To validate this co-design process, we conducted a targeted ablation study using our best-performing algorithm, BRKGA+Llama-4-Maverick. The objective is to establish the soundness of our metric selection methodology by addressing two key questions: (1) Is using four metrics a reasonable compromise between performance and complexity compared to a smaller set? and (2) What is the true value of human curation compared to a random selection of LLM-suggested metrics?

\paragraph{Case 1: Performance with a Reduced Metric Set.}
For the first test, we evaluated a simplified version of our framework using only the two primary metrics identified in Section~\ref{subsec:phase_2}: Normalized Length ($M_L$) and Opportunity ($M_O$). As shown in Table~\ref{tab:ablation_metrics} (Case 1), the performance with only two metrics is surprisingly strong, achieving an overall average score of 312.88. While this is slightly below our full framework's score (313.07), it is notably better than the other LLM-hybrids that used all four metrics (see Table~\ref{tab:brkga_vs_llm}). This finding is significant: it indicates that Llama-4-Maverick can achieve highly competitive results with a simpler heuristic, but the addition of two more curated metrics allows it to reach a higher performance ceiling, particularly on the most complex instances. This underscores the value of the human-curated feature set in achieving state-of-the-art performance.

\paragraph{Case 2: Performance with a Randomly Selected Metric Set.}
For the second test, we replaced the human curation in Phase 1 with a random process, selecting four metrics at random from the ten candidates proposed by the LLM. The results in Table~\ref{tab:ablation_metrics} (Case 2) reveal a surprising outcome: the random set is highly competitive, achieving an average score (312.94) nearly identical to our curated framework's (313.07) and converging faster on average (78.51s vs. 89.70s). This finding initially suggests that human curation might be superfluous for identifying an effective heuristic signal.

\begin{figure}[ht]
    \centering
    \includegraphics[width=0.7\linewidth]{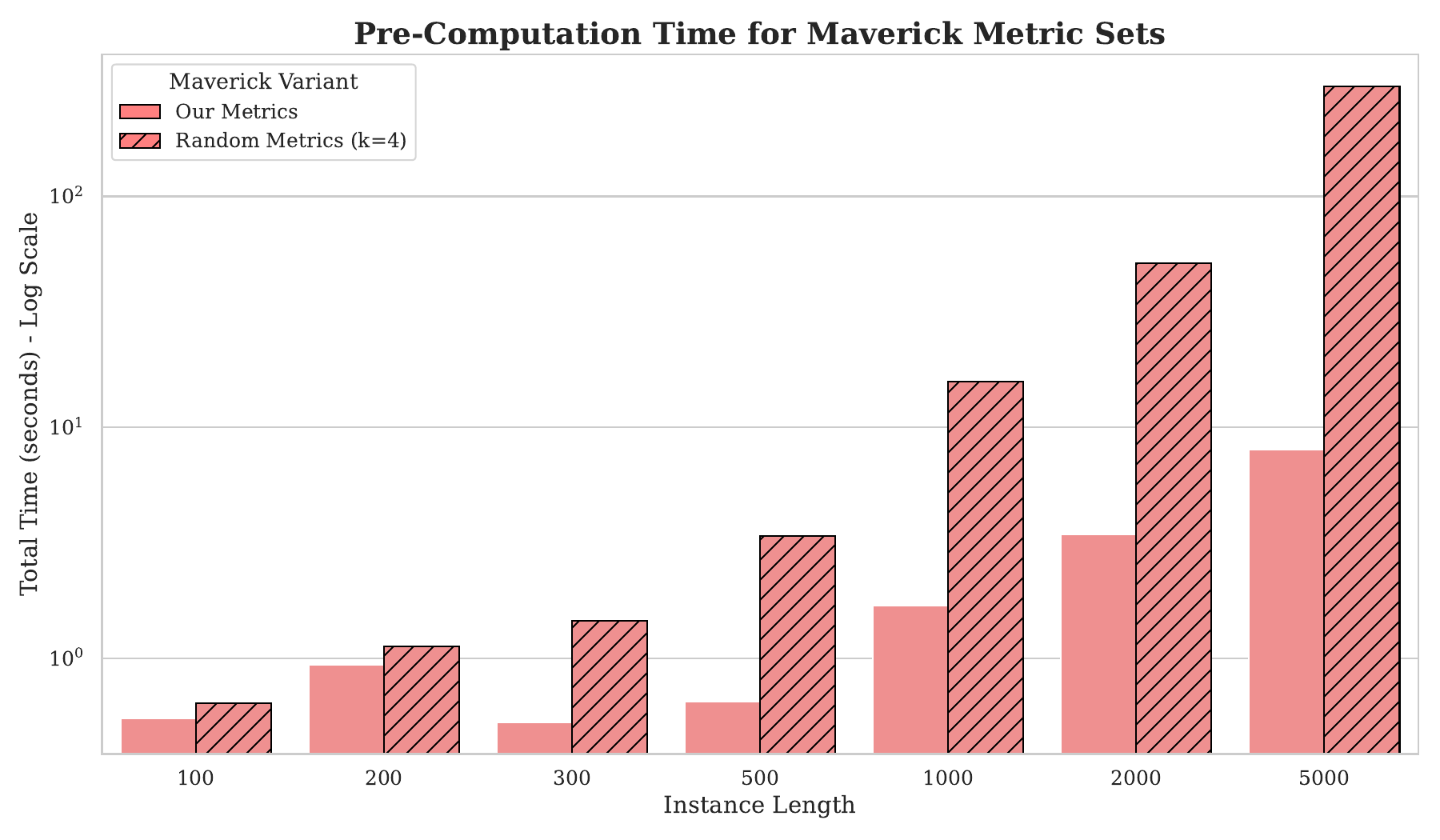}
\caption{Pre-computation time comparison for metric sets. The analysis contrasts the total time required to compute our four curated metrics (solid bar) against a set of four randomly selected metrics (hatched bar).}
    \label{fig:comparison_metrics_time}
\end{figure}

However, such a conclusion would overlook a critical dimension of performance: the \textit{pre-computation cost} of the metrics themselves. As shown in Figure~\ref{fig:comparison_metrics_time}, the computational overhead of the two sets differs dramatically. While the calculation of the curated metric set only took 8.01 seconds for instances with input string length $n=5000$, the calculation of the four randomly selected metrics required 300.06 seconds---more than 17 times longer. This reveals the true value of the human-in-the-loop process: it acts as an essential filter for \textit{computational efficiency}. While the LLM is adept at generating heuristically effective metrics, it lacks the judgment to assess their algorithmic complexity. The human expert's role is therefore fundamental, not necessarily to find a marginally better heuristic, but to ensure the final set is practically viable and computationally tractable. This insight opens avenues for future work where LLMs could be trained to evaluate metrics based on both their heuristic value and their computational cost.

For one of our test runs, the four randomly selected metrics proposed by the LLM, whose complexity explains their high pre-computation time, were:
\begin{itemize}
    \item \textbf{Character Change Frequency:} For each run, calculate the number of subsequent character changes in the string $S$, normalized.
    \item \textbf{Sequence Break Potential:} For each run, calculate the proportion of other runs that would become unreachable if this run were to be selected.
    \item \textbf{Immediate Next Run Length:} The raw length of the very next run in the sequence.
    \item \textbf{External Fragmentation Potential:} As defined in Section~\ref{subsec:phase_2}.
\end{itemize}

\begin{table}
    \centering
  \caption{Ablation study for \textbf{BRKGA+Llama-4-Maverick}, our best-performing variant from Table~\ref{tab:brkga_vs_llm}. The analysis compares the performance of our framework's curated 4-metric set against a simpler 2-metric set and a randomly chosen 4-metric set. The best average solution (`avg.') is highlighted, and the fastest time (`time') is marked with a {\scriptsize\textcolor{orange}{\faBolt}} icon.}
    \label{tab:ablation_metrics}
    
    \sisetup{
        output-decimal-marker={.},
        group-separator={,},
        table-format=4.2
    }

    \begin{tabular}{
        c c 
        S S[table-format=3.2] 
        S S[table-format=3.2] 
        S S[table-format=3.2]
    }
        \toprule
        & & \multicolumn{2}{c}{\textbf{Case 1}} & \multicolumn{2}{c}{\textbf{Case 2}} & \multicolumn{2}{c}{\textbf{Our Metrics}} \\
        & & \multicolumn{2}{c}{($k=2$ Simple)} & \multicolumn{2}{c}{($k=4$ Random)} & \multicolumn{2}{c}{($k=4$ Curated)} \\
        \cmidrule(lr){3-4} \cmidrule(lr){5-6} \cmidrule(lr){7-8}
        \textbf{Length} & \textbf{$\Sigma$} & {\textbf{avg.}} & {\textbf{time}} & {\textbf{avg.}} & {\textbf{time}} & {\textbf{avg.}} & {\textbf{time}} \\
        \midrule
        \multirow{5}{*}{100} 
        & 2 & \bestavg{59.03} & 0.00\fastest & \bestavg{59.03} & 0.00\fastest & \bestavg{59.03} & 0.00\fastest \\
        & 4 & \bestavg{41.07} & 0.01\fastest & \bestavg{41.07} & 0.02 & \bestavg{41.07} & 0.02 \\
        & 8 & 33.83 & 0.71\fastest & 33.87 & 1.16 & \bestavg{33.90} & 0.92 \\
        & 16 & 34.57 & 1.96\fastest & 34.20 & 3.25 & \bestavg{34.60} & 3.04 \\
        & 32 & \bestavg{41.13} & 9.14 & 39.47 & 7.60\fastest & 41.10 & 8.29 \\
        \addlinespace
        \multirow{5}{*}{200} 
        & 2 & \bestavg{115.17} & 0.00\fastest & \bestavg{115.17} & 0.00\fastest & \bestavg{115.17} & 0.00\fastest \\
        & 4 & \bestavg{72.77} & 0.12 & \bestavg{72.77} & 0.09\fastest & \bestavg{72.77} & 0.17 \\
        & 8 & \bestavg{55.17} & 2.40 & \bestavg{55.17} & 3.24 & 55.13 & 1.87\fastest \\
        & 16 & 50.90 & 11.10\fastest & 50.43 & 13.99 & \bestavg{51.03} & 11.30 \\
        & 32 & 55.30 & 23.13 & 53.83 & 22.13 & \bestavg{55.87} & 21.21\fastest \\
        \addlinespace
        \multirow{5}{*}{300} 
        & 2 & \bestavg{165.47} & 0.00\fastest & \bestavg{165.47} & 0.00\fastest & \bestavg{165.47} & 0.00\fastest \\
        & 4 & \bestavg{102.73} & 0.33 & \bestavg{102.73} & 0.25\fastest & \bestavg{102.73} & 0.74 \\
        & 8 & 75.07 & 2.54\fastest & 74.93 & 5.01 & \bestavg{75.10} & 4.17 \\
        & 16 & \bestavg{66.63} & 18.39\fastest & 66.40 & 20.40 & \bestavg{66.63} & 21.37 \\
        & 32 & 67.27 & 33.90 & 67.50 & 34.37 & \bestavg{67.63} & 33.17\fastest \\
        \addlinespace
        \multirow{5}{*}{500} 
        & 2 & \bestavg{271.67} & 0.01\fastest & \bestavg{271.67} & 0.01\fastest & \bestavg{271.67} & 0.02 \\
        & 4 & \bestavg{162.77} & 0.73 & \bestavg{162.77} & 0.53 & \bestavg{162.77} & 0.45\fastest \\
        & 8 & 110.07 & 13.92 & 110.17 & 13.02\fastest & \bestavg{110.20} & 21.53 \\
        & 16 & 91.17 & 43.03 & \bestavg{91.27} & 32.13\fastest & 91.17 & 42.28 \\
        & 32 & 86.57 & 61.45 & \bestavg{88.63} & 51.85\fastest & 87.17 & 63.28 \\
        \addlinespace
        \multirow{5}{*}{1000} 
        & 2 & \bestavg{530.23} & 0.03 & \bestavg{530.23} & 0.03 & \bestavg{530.23} & 0.02\fastest \\
        & 4 & \bestavg{301.10} & 5.43 & \bestavg{301.10} & 0.86\fastest & 301.00 & 6.18 \\
        & 8 & \bestavg{192.03} & 41.42 & 191.87 & 27.32\fastest & 191.80 & 34.84 \\
        & 16 & 143.80 & 101.11 & \bestavg{144.70} & 77.14\fastest & 143.37 & 107.15 \\
        & 32 & 125.60 & 128.71 & \bestavg{130.30} & 131.77 & 126.23 & 140.86 \\
        \addlinespace
        \multirow{5}{*}{2000} 
        & 2 & \bestavg{1041.73} & 0.14 & \bestavg{1041.73} & 0.44 & \bestavg{1041.73} & 0.12\fastest \\
        & 4 & 572.57 & 28.39 & \bestavg{572.67} & 24.35\fastest & 572.50 & 29.98 \\
        & 8 & 343.43 & 145.29 & 343.70 & 83.45 & \bestavg{343.90} & 65.82\fastest \\
        & 16 & 234.87 & 239.00 & \bestavg{236.83} & 179.00\fastest & 235.10 & 199.04 \\
        & 32 & 188.97 & 309.65 & 188.03 & 259.36\fastest & \bestavg{189.90} & 317.78 \\
        \addlinespace
        \multirow{5}{*}{5000} 
        & 2 & \bestavg{2567.47} & 0.41 & \bestavg{2567.47} & 1.49 &  \bestavg{2567.47} & 0.22\fastest \\
        & 4 & \bestavg{1361.33} & 49.68\fastest & 1361.07 & 53.18 & 1361.17 & 78.86 \\
        & 8 & \bestavg{768.67} & 314.58\fastest & 766.63 & 339.24 & 768.07 & 372.71 \\
        & 16 & 482.90 & 687.23 & 482.90 & 581.52\fastest & \bestavg{486.23} & 742.38 \\
        & 32 & 337.90 & 845.29 & 337.13 & 779.63\fastest & \bestavg{338.40} & 809.65 \\
        \midrule
        \textbf{Average} & & 312.88 & 89.12 & 312.94 & 78.51\fastest & \bestavg{313.07} & 89.70 \\
        \bottomrule
    \end{tabular}
\end{table}

\subsection{Validating the Heuristic Contribution of the LLM}\label{sec:valid_phase_3}

Given that BRKGA is known to be susceptible to external bias, a critical question arises: is the guidance from the LLM genuinely heuristic or merely equivalent to random noise? To answer this, we conducted an ablation study (Table~\ref{tab:final_comparison_en_reordered}) benchmarking our LLM-guided approach against two baselines that use randomly generated bias vectors:
\begin{itemize}
    \item \textbf{Static Randomness:} The bias vector, $\vec{L}_{\scriptstyle\text{random}}^{\scriptstyle\text{static}}$, is generated once from a uniform random distribution at the start of the execution and remains fixed throughout all generations.
    \item \textbf{Dynamic Randomness:} The bias vector, $\vec{L}_{\scriptstyle\text{random}}^{\scriptstyle\text{dynamic}}$, is regenerated with new uniform random values at the beginning of each population generation.
\end{itemize}

It is important to recall that each element in the vector $\vec{L}$ corresponds to a specific run of the considered LRS instance $S$ (see Section~\ref{subsec:phase_4}). In our framework, this vector is constructed from the alpha-beta values returned by the LLM after it analyzes the prompt containing the tabular metric data, as detailed in Section~\ref{subsec:phase_3}.

The results show that the LLM-guided approach consistently and significantly outperforms both random baselines. The only exceptions occur for the simplest instances ($|\Sigma| \in \{2, 4\}$), where performance is statistically tied---a finding consistent with our previous analysis. Interestingly, the data also suggests that the dynamic random vector tends to outperform its static counterpart. This makes the superiority of our own static, LLM-generated vector even more noteworthy. The fact that our single, pre-computed $\vec{L}$ vector is more effective than a bias vector that is randomly regenerated at every generation provides strong evidence that the LLM's guidance is not arbitrary. Instead, it is providing a coherent and effective heuristic strategy based on the instance's specific structure, which becomes increasingly crucial as the problem's complexity grows.

Perhaps the most striking finding is that, when the random baselines ($\vec{L}_{\scriptstyle\text{random}}^{\scriptstyle\text{static}}$ and $\vec{L}_{\scriptstyle\text{random}}^{\scriptstyle\text{dynamic}}$) are compared with the standard BRKGA in Table~\ref{tab:brkga_vs_llm}, their performance is actually worse (see row averages). The average value of each random vector for every LLM is below 310, indicating the presence of a \textit{negative bias} that drives the BRKGA towards less promising search spaces. This suggests that introducing an arbitrary, random bias is not only inferior to the intelligent guidance provided by our framework, but can in fact be detrimental to the search, yielding results worse than the baseline without any external information.

\begin{sidewaystable}
    \centering
    \caption{Performance of LLM-integrated BRKGAs, in comparison to those BRKGA variants in which the vector $\vec{L}$ is generated once before execution (static) or dynamically chosen during runtime (dynamic). This is to verify that the gains from LLM-generated probabilities are not merely by chance. The best-performing values in each row are highlighted in light yellow.}
    \label{tab:final_comparison_en_reordered}
    \scalebox{0.8}{
    \begin{tabular}{cc|ccc|ccc|ccc|ccc}
        \toprule
        \textbf{Length} & $\Sigma$ & \multicolumn{3}{c}{\textbf{GPT-4.1-mini}} & \multicolumn{3}{c}{\textbf{Gemini-2.5-Flash}} & \multicolumn{3}{c}{\textbf{Llama-3.2-3b}} & \multicolumn{3}{c}{\textbf{Llama-4-Maverick}} \\
        \cmidrule(lr){3-5} \cmidrule(lr){6-8} \cmidrule(lr){9-11} \cmidrule(lr){12-14}
        & & LLM & Rand. Static & Rand. Dynamic & LLM & Rand. Static & Rand. Dynamic & LLM & Rand. Static & Rand. Dynamic & LLM & Rand. Static & Rand. Dynamic \\
        \midrule
        \multirow{5}{*}{100} 
        & 2 & \cellcolor{lightyellow}{\textbf{59.03}} & 58.93 & \cellcolor{lightyellow}{\textbf{59.03}} & \cellcolor{lightyellow}{\textbf{59.03}} & 58.93 & \cellcolor{lightyellow}{\textbf{59.03}} & \cellcolor{lightyellow}{\textbf{59.03}} & 58.93 & \cellcolor{lightyellow}{\textbf{59.03}} & \cellcolor{lightyellow}{\textbf{59.03}} & 58.93 & 58.93 \\
        & 4 & \cellcolor{lightyellow}{\textbf{41.07}} & 40.80 & 40.83 & \cellcolor{lightyellow}{\textbf{41.07}} & 40.87 & 40.83 & \cellcolor{lightyellow}{\textbf{41.07}} & 40.87 & 40.60 & \cellcolor{lightyellow}{\textbf{41.07}} & 40.87 & \cellcolor{lightyellow}{\textbf{41.07}} \\
        & 8 & \cellcolor{lightyellow}{\textbf{33.90}} & 33.13 & 32.70 & 33.87 & 33.00 & 32.60 & 33.87 & 33.03 & 32.97 & \cellcolor{lightyellow}{\textbf{33.90}} & 33.27 & 32.83 \\
        & 16 & 34.57 & 31.70 & 31.53 & 34.57 & 32.07 & 31.23 & 34.53 & 32.10 & 32.57 & \cellcolor{lightyellow}{\textbf{34.60}} & 32.43 & 32.10 \\
        & 32 & 40.40 & 34.10 & 33.43 & 40.93 & 34.93 & 34.33 & 40.83 & 34.30 & 34.40 & \cellcolor{lightyellow}{\textbf{41.10}} & 35.60 & 35.10 \\
        \midrule
        \multirow{5}{*}{200} 
        & 2 & \cellcolor{lightyellow}{\textbf{115.17}} & 115.10 & \cellcolor{lightyellow}{\textbf{115.17}} & \cellcolor{lightyellow}{\textbf{115.17}} & 115.10 & 115.00 & \cellcolor{lightyellow}{\textbf{115.17}} & 115.10 & 115.13 & \cellcolor{lightyellow}{\textbf{115.17}} & 115.10 & 115.03 \\
        & 4 & \cellcolor{lightyellow}{\textbf{72.77}} & 72.50 & 72.40 & \cellcolor{lightyellow}{\textbf{72.77}} & 72.57 & 72.30 & \cellcolor{lightyellow}{\textbf{72.77}} & 72.53 & 72.53 & \cellcolor{lightyellow}{\textbf{72.77}} & 72.67 & 72.53 \\
        & 8 & 55.10 & 54.07 & 53.90 & \cellcolor{lightyellow}{\textbf{55.17}} & 54.20 & 54.07 & 55.10 & 54.10 & 54.00 & 55.13 & 54.13 & 53.93 \\
        & 16 & 50.83 & 46.47 & 47.03 & 50.87 & 47.17 & 47.23 & 50.83 & 46.63 & 47.83 & \cellcolor{lightyellow}{\textbf{51.03}} & 47.37 & 48.40 \\
        & 32 & 55.37 & 46.40 & 47.20 & 54.50 & 47.17 & 45.70 & 55.13 & 46.23 & 47.20 & \cellcolor{lightyellow}{\textbf{55.87}} & 47.77 & 47.37 \\
        \midrule
        \multirow{5}{*}{300} 
        & 2 & \cellcolor{lightyellow}{\textbf{165.47}} & 165.27 & 165.43 & \cellcolor{lightyellow}{\textbf{165.47}} & 165.43 & 165.33 & \cellcolor{lightyellow}{\textbf{165.47}} & 165.27 & 165.37 & \cellcolor{lightyellow}{\textbf{165.47}} & 165.43 & \cellcolor{lightyellow}{\textbf{165.47}} \\
        & 4 & \cellcolor{lightyellow}{\textbf{102.73}} & 102.43 & 102.50 & \cellcolor{lightyellow}{\textbf{102.73}} & 102.50 & 102.40 & \cellcolor{lightyellow}{\textbf{102.73}} & 102.53 & 102.53 & \cellcolor{lightyellow}{\textbf{102.73}} & 102.53 & 102.47 \\
        & 8 & 75.07 & 73.30 & 73.57 & \cellcolor{lightyellow}{\textbf{75.10}} & 73.67 & 73.47 & 75.07 & 73.33 & 73.70 & \cellcolor{lightyellow}{\textbf{75.10}} & 73.77 & 73.60 \\
        & 16 & 66.30 & 61.40 & 61.87 & \cellcolor{lightyellow}{\textbf{66.77}} & 61.93 & 62.00 & 66.27 & 61.90 & 62.87 & 66.63 & 62.47 & 62.90 \\
        & 32 & 66.03 & 57.00 & 56.90 & 66.50 & 58.07 & 56.90 & 66.47 & 57.77 & 58.13 & \cellcolor{lightyellow}{\textbf{67.63}} & 59.23 & 58.93 \\
        \midrule
        \multirow{5}{*}{500} 
        & 2 & \cellcolor{lightyellow}{\textbf{271.67}} & 271.50 & \cellcolor{lightyellow}{\textbf{271.67}} & \cellcolor{lightyellow}{\textbf{271.67}} & 271.50 & 271.60 & \cellcolor{lightyellow}{\textbf{271.67}} & 271.50 & 271.63 & \cellcolor{lightyellow}{\textbf{271.67}} & 271.50 & \cellcolor{lightyellow}{\textbf{271.67}} \\
        & 4 & 162.73 & 162.33 & 162.33 & \cellcolor{lightyellow}{\textbf{162.77}} & 162.33 & 162.47 & \cellcolor{lightyellow}{\textbf{162.77}} & 162.30 & 162.30 & \cellcolor{lightyellow}{\textbf{162.77}} & 162.43 & 162.67 \\
        & 8 & 109.93 & 108.40 & 108.27 & \cellcolor{lightyellow}{\textbf{110.20}} & 108.60 & 108.40 & 110.13 & 108.57 & 108.37 & \cellcolor{lightyellow}{\textbf{110.20}} & 108.70 & 108.67 \\
        & 16 & 90.60 & 85.30 & 85.40 & 90.80 & 85.07 & 85.13 & 90.33 & 86.17 & 86.40 & \cellcolor{lightyellow}{\textbf{91.17}} & 86.53 & 87.33 \\
        & 32 & 85.50 & 74.43 & 74.70 & 85.77 & 75.37 & 74.50 & 86.07 & 74.97 & 75.47 & \cellcolor{lightyellow}{\textbf{87.17}} & 77.27 & 75.70 \\
        \midrule
        \multirow{5}{*}{1000} 
        & 2 & \cellcolor{lightyellow}{\textbf{530.23}} & 530.03 & 530.07 & \cellcolor{lightyellow}{\textbf{530.23}} & 530.03 & 530.17 & \cellcolor{lightyellow}{\textbf{530.23}} & 530.03 & 530.10 & \cellcolor{lightyellow}{\textbf{530.23}} & 530.03 & 530.10 \\
        & 4 & \cellcolor{lightyellow}{\textbf{301.10}} & 300.53 & 300.27 & \cellcolor{lightyellow}{\textbf{301.10}} & 300.50 & 300.60 & 301.00 & 300.37 & 300.80 & 301.00 & 300.60 & 301.03 \\
        & 8 & \cellcolor{lightyellow}{\textbf{191.87}} & 189.57 & 189.33 & 191.23 & 189.73 & 188.97 & 191.23 & 189.60 & 189.53 & 191.80 & 189.90 & 189.87 \\
        & 16 & 142.90 & 137.27 & 137.63 & 143.33 & 138.17 & 136.83 & 142.77 & 137.53 & 137.80 & \cellcolor{lightyellow}{\textbf{143.37}} & 138.13 & 138.43 \\
        & 32 & 123.97 & 111.43 & 110.60 & 124.00 & 112.33 & 108.47 & 123.97 & 110.47 & 111.73 & \cellcolor{lightyellow}{\textbf{126.23}} & 113.20 & 110.90 \\
        \midrule
        \multirow{5}{*}{2000} 
        & 2 & \cellcolor{lightyellow}{\textbf{1041.73}} & 1041.70 & 1041.57 & \cellcolor{lightyellow}{\textbf{1041.73}} & 1041.70 & 1041.67 & \cellcolor{lightyellow}{\textbf{1041.73}} & 1041.63 & 1041.70 & \cellcolor{lightyellow}{\textbf{1041.73}} & 1041.70 & 1041.70 \\
        & 4 & \cellcolor{lightyellow}{\textbf{572.67}} & 572.13 & 572.03 & 572.50 & 572.07 & 572.07 & 572.50 & 572.03 & 572.27 & 572.50 & 572.33 & 572.30 \\
        & 8 & 343.17 & 340.63 & 340.90 & 343.10 & 340.87 & 339.87 & 343.70 & 340.83 & 340.73 & \cellcolor{lightyellow}{\textbf{343.90}} & 340.50 & 342.23 \\
        & 16 & 234.63 & 227.13 & 224.63 & 235.00 & 227.13 & 226.03 & 233.73 & 226.97 & 227.23 & \cellcolor{lightyellow}{\textbf{235.10}} & 231.37 & 229.07 \\
        & 32 & 186.80 & 167.77 & 168.67 & 188.13 & 167.53 & 166.57 & 186.23 & 167.97 & 168.73 & \cellcolor{lightyellow}{\textbf{189.90}} & 171.17 & 170.80 \\
        \midrule
        \multirow{5}{*}{5000} 
        & 2 & \cellcolor{lightyellow}{\textbf{2567.47}} & \cellcolor{lightyellow}{\textbf{2567.47}} & \cellcolor{lightyellow}{\textbf{2567.47}} & \cellcolor{lightyellow}{\textbf{2567.47}} & \cellcolor{lightyellow}{\textbf{2567.47}} & 2567.33 & \cellcolor{lightyellow}{\textbf{2567.47}} & \cellcolor{lightyellow}{\textbf{2567.47}} & 2567.30 & \cellcolor{lightyellow}{\textbf{2567.47}} & \cellcolor{lightyellow}{\textbf{2567.47}} & 2567.37 \\
        & 4 & \cellcolor{lightyellow}{\textbf{1361.30}} & 1360.70 & 1360.07 & 1361.17 & 1360.50 & 1360.80 & \cellcolor{lightyellow}{\textbf{1361.30}} & 1360.47 & 1360.57 & 1361.17 & 1361.17 & 1360.67 \\
        & 8 & 766.87 & 762.83 & 763.40 & \cellcolor{lightyellow}{\textbf{768.50}} & 762.53 & 760.27 & 767.13 & 762.43 & 765.53 & 768.07 & 762.77 & 765.00 \\
        & 16 & 483.57 & 470.90 & 473.83 & 481.33 & 471.67 & 471.53 & 483.90 & 468.60 & 473.97 & \cellcolor{lightyellow}{\textbf{486.23}} & 474.97 & 477.23 \\
        & 32 & 335.73 & 315.80 & 312.67 & 336.93 & 313.40 & 318.87 & 336.50 & 315.03 & 312.73 & \cellcolor{lightyellow}{\textbf{338.40}} & 321.80 & 316.87 \\
        \midrule
        \multicolumn{2}{l}{\textbf{Average}} 
        & 312.52 & 308.30 & 308.26 & 312.61 & 308.46 & 308.13 & 312.53 & 308.27 & 308.69 & \cellcolor{lightyellow}{\textbf{313.07}} & 309.29 & 309.15 \\
        \bottomrule
    \end{tabular}
    }
\end{sidewaystable}
\subsection{Analysis of Algorithmic Behavior}\label{sec:analysis-behavior}

While the numerical and statistical results confirm that the BRKGA+LLM hybrids are superior, a key question remains regarding the qualitative differences in their \textit{search behavior}. To answer this, we employ the STNWeb~\footnote{See \url{https://stn-analytics.com/} and \url{https://github.com/camilochs/stnweb}.} visualization tool~\cite{CHACONSARTORI2023100558}, which models the optimization process as a Search Trajectory Network (STN)~\cite{OCHOA2021107492}. Our hypothesis is that the standard BRKGA will tend to converge prematurely into a few large \textit{basins of attraction}, whereas the LLM-guided hybrid should exhibit a more diverse and effective exploration of the solution space.

\subsubsection{STN terminology}

Figure~\ref{fig:stnweb-analysis} presents the STNs generated by comparing the baseline BRKGA against our best-performing hybrid, BRKGA+Llama-4-Maverick. To interpret these diagrams, the visual terminology of the STN is defined as follows:
\begin{itemize}
    \item Each algorithm run (trajectory) originates at a yellow square (\tikz[baseline=-0.5ex] \node[draw=yellow!50!black,fill=yellow,minimum size=3mm,inner sep=0pt,shape=rectangle] {};) 
    and terminates either at a black, right-pointing triangle (\tikz[baseline=-0.5ex] \node[draw=black,fill=black,isosceles triangle,isosceles triangle apex angle=60,shape border rotate=90,minimum size=3mm,inner sep=0pt] {};), or at a red dot (\tikz[baseline=-0.5ex] \node[draw=red,fill=red,minimum size=3mm,inner sep=0pt,shape=circle] {};) in case the trajectory endpoint corresponds to a best solution found in the comparison.
    
    \item Any node of the STN diagram shows a chunk of the search space that contains, in the simplest case, a single solution. However, when plotting such complete STN networks, it is often difficult to see the main characteristics of the search process. Therefore, STNWeb provides multiple methods for clustering similar solutions into single nodes of the STN diagram. This is called \emph{search space partitioning} in STN terminology. The STN graphics in Figure~\ref{fig:stnweb-analysis} are produced with agglomerative clustering for search space partitioning, with parameters \texttt{cluster size} and \texttt{volume size} at 5\%, and utilizing the default Hamming distance metric for measuring distances. Moreover, for the five STN-graphics in Figure~\ref{fig:stnweb-analysis}, corresponding to $|\Sigma| \in \{2, 4, 8, 16, 32\}$, the number of clusters was set to 134, 280, 255, 211, and 260, respectively. These values were chosen to provide the clearest possible visualization for each level of instance complexity.
        
    \item The size of a node is proportional to the number of trajectories that converged to any of the solutions contained in this node. Gray nodes (\tikz[baseline=-0.5ex] \node[draw=gray,fill=gray,minimum size=3mm,inner sep=0pt,shape=circle] {};) 
    indicate solutions found by \textit{both} the baseline and the hybrid algorithm, highlighting shared regions of the search space.

    \item Edges in STN diagrams connect the nodes to show the step-by-step path of the algorithms' search trajectories.
\end{itemize}

The figure visually confirms our hypothesis. At low alphabet sizes ($|\Sigma|=2$), both algorithms explore a simple search space in which best solutions are rather easy to find. However, as complexity increases, the trajectories of the baseline BRKGA (colored purple \tikz[baseline=-0.5ex] \node[draw=purple,fill=purple,minimum size=3mm,inner sep=0pt,shape=circle] {};) tend to converge into a single, dense basin of attraction, indicating premature convergence. In contrast, the BRKGA+Llama-4-Maverick trajectories (colored light red \tikz[baseline=-0.5ex] \node[draw={rgb,255:red,255;green,128;blue,128},fill={rgb,255:red,255;green,128;blue,128},minimum size=3mm,inner sep=0pt,shape=circle] {};) remain exploratory, covering a much wider region of the solution space.

\subsubsection{STN analyzing the LRS}

\begin{figure}
    \centering
    \includegraphics[width=0.8\linewidth]{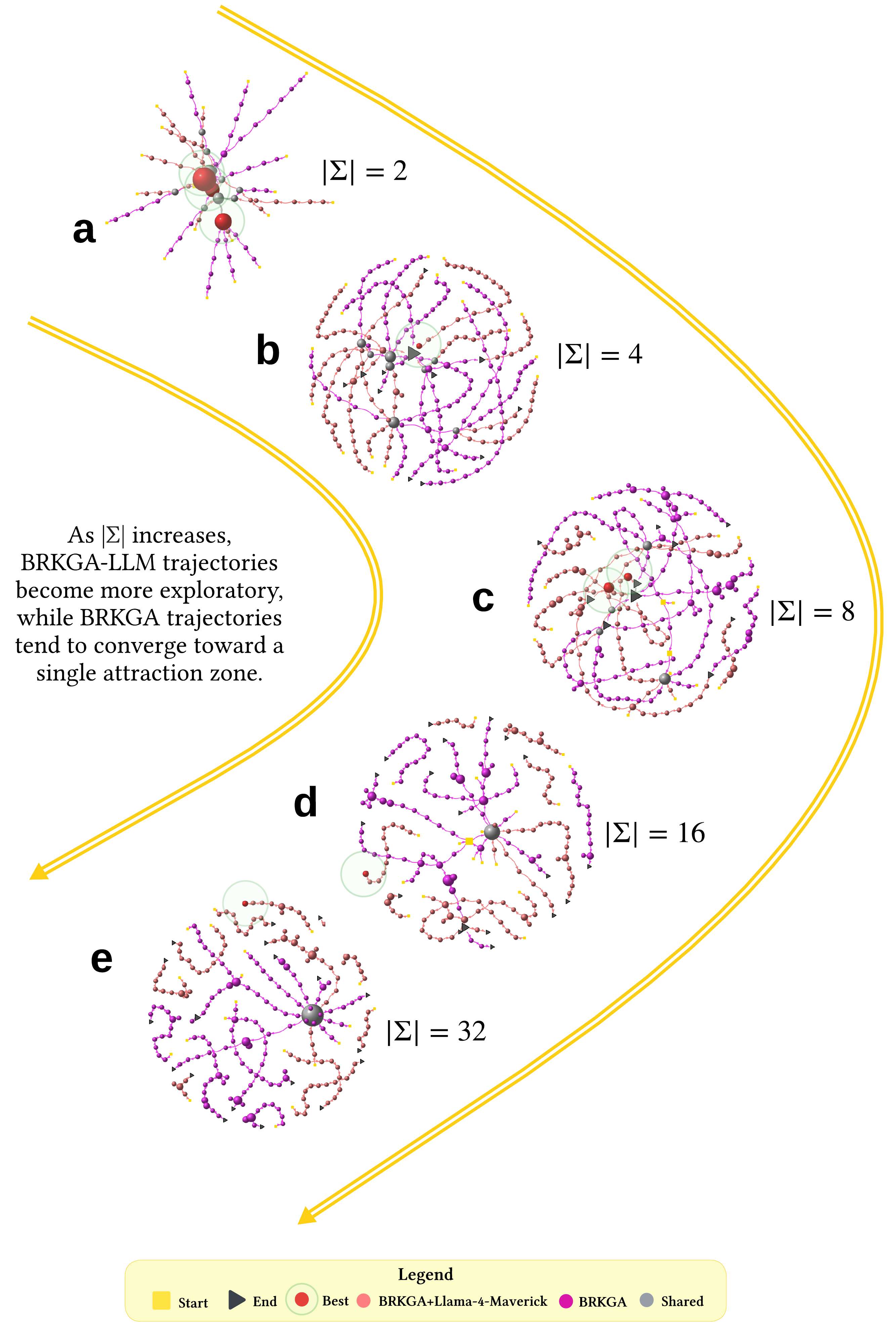}
    \caption{Five images generated with STNWeb~\cite{CHACONSARTORI2023100558} comparing the search behavior of BRKGA+Llama-4-Maverick and the standard BRKGA on an instance with $n=5000$ across varying alphabet sizes ($|\Sigma|$). (a) For a small alphabet ($|\Sigma|=2$), both algorithms initially locate a best solution (red nodes). However, as $|\Sigma|$ increases, BRKGA+Llama-4-Maverick exhibits enhanced exploration capabilities, while the standard BRKGA tends to converge prematurely to a single basin of attraction, thus failing to find better solutions (e). This indicates that the guidance provided by the LLM is neither arbitrary nor random, but rather effectively steers the algorithm toward more promising regions of the search space.}

    \label{fig:stnweb-analysis}
\end{figure}

We now analyze the STN for each level of instance complexity, defined by the alphabet size $|\Sigma|$, as shown in Figure~\ref{fig:stnweb-analysis}. Each trajectory represents an attempt to solve the LRS problem on an instance with input stringn length $n=5000$.

\begin{enumerate}[label=\alph*)]
    \item For $|\Sigma|=2$, the LRS problem is highly structured with long, easily identifiable runs. As expected, the search trajectories of both algorithms are largely overlapping, and both consistently locate best-found solutions (red nodes). This visual finding aligns with the numerical results in Table~\ref{tab:brkga_vs_llm}, where the performances were statistically indistinguishable.

    \item For $|\Sigma|=4$, the increased alphabet size introduces more combinatorial possibilities, causing the search paths not to converge so easily. While both algorithms remain exploratory and do not get trapped, we see the emergence of more shared solutions (gray nodes) at intermediate stages of the search process.

    \item For $|\Sigma|=8$, the LRS problem becomes significantly more challenging. The STN network grows denser as both algorithms find many of the same intermediate-quality solutions. However, the search remains well-distributed, indicating that there are still multiple viable pathways to good solutions, thus preventing premature convergence.

    \item For $|\Sigma|=16$, a critical shift occurs. The input string now exhibits sufficient combinatorial complexity that the baseline BRKGA trajectories begin to collapse toward a central basin of attraction, repeatedly becoming trapped in the same locally optimal region. In contrast, while some of the BRKGA+Llama-4-Maverick's trajectories explore this shared region, many others remain independent, demonstrating that the LLM's guidance helps it maintain a more diverse and effective search.
    
    \item This tendency is reinforced for $|\Sigma|=32$. At this alphabet size, the search space of an LRS instance becomes highly complex. The standard BRKGA is now rather ineffective, with the vast majority of its runs converging to a single suboptimal solution. BRKGA+Llama-4-Maverick, however, avoids this trap almost entirely. Its ability to continue exploring diverse regions and ultimately find better solutions demonstrates that the LLM's data-driven guidance is most crucial when the problem's inherent structure is least apparent.
\end{enumerate}

This behavioral analysis provides the visual evidence for our central hypothesis. While the preceding statistical tests confirm \textit{that} an algorithm performs better on average, the STN visualization in Figure~\ref{fig:stnweb-analysis} reveals \textit{how} this is achieved. The diagram demonstrates that the LLM's guidance fundamentally alters the search dynamics, preventing the premature convergence seen in the baseline and promoting a more robust exploration of the solution space---a crucial insight that aggregate numerical data alone cannot provide.

\subsection{Analysis of LLM-Generated Parameters}\label{sec:analysis-llm-parameters}

To understand the heuristic strategies developed by each LLM, this section provides an in-depth analysis of the alpha ($\alpha$) and beta ($\beta$) parameters they generate. This investigation is a novel contribution, as the foundational work in~\cite{10818476} did not examine the emergent decision-making process of the models. 

Our analysis is presented in Figure~\ref{fig:alpha_beta_analysis} using violin plots, selected over standard box plots for their ability to convey the complete probability distribution of the data rather than only summary statistics such as median and interquartile range. By combining a box plot with a kernel density estimation, violin plots reveal important distributional features---including shape, skewness, multimodality, and variability---that would otherwise remain hidden. This is particularly relevant in our context, where subtle changes in the $\alpha$ and $\beta$ parameters may indicate distinct heuristic behaviors across LLMs, even when central tendencies appear similar.

The graphic is organized into four subplots comparing instances with $n = 100$ and $n = 5000$. In each subplot, the rows correspond to $|\Sigma| \in \{2, 4, 8, 16, 32\}$, and the columns represent the parameters. For each combination, 30 random instances were generated, resulting in a total of $1050$ instances. The eight parameters shown correspond, in order, to the four metrics defined in Section~\ref{subsec:phase_2}: Normalized Length ($M_L$), Opportunity ($M_O$), Distance to Next Run ($M_D$), and Global Character Frequency ($M_F$).

\subsubsection{Analysis of Alpha Parameter Distributions}

The alpha parameters ($\alpha_i$) denote the \textit{relative importance} or \textit{weight} assigned by the LLM to each of the four heuristic metrics, with the constraint that they sum to one ($\sum_{i=1}^{4} \alpha_i = 1$). A close examination of their distributions, especially for the top-performing BRKGA+Llama-4-Maverick, uncovers several important insights into the utilized heuristic strategies.

We now turn to interpreting the findings for the $\alpha_i$ parameters presented in the first column---subplots (a) and (c) of Figure~\ref{fig:alpha_beta_analysis}.

\paragraph{Consensus on Primary Metrics.}
A clear consensus emerges for the primary metrics, $\alpha_1$ (Normalized Length) and $\alpha_2$ (Opportunity), especially for large instances ($n=5000$). For these parameters, most LLM variants, including Llama-4-Maverick, exhibit very low variance, producing tall, narrow violin plots. This indicates a strong, unanimous agreement that these two metrics are fundamentally the most important for guiding the search. The models confidently and consistently assign them high weights, confirming they have correctly identified the main drivers of solution quality in the LRS problem.

\paragraph{Divergence and Adaptability in Secondary Metrics.}
In contrast, the strategies for weighting the secondary metrics, $\alpha_3$ (Distance) and $\alpha_4$ (Frequency), are highly model-dependent and reveal Llama-4-Maverick's unique adaptability. While other LLMs like Gemini-2.5-Flash and GPT-4.1-mini produce highly consistent, low-variance weights for these metrics, Llama-4-Maverick displays significant variability. This suggests that Llama-4-Maverick is dynamically adjusting the importance of these secondary factors based on patterns it detects in each specific instance.

This adaptive behavior is best illustrated by Llama-4-Maverick's treatment of $\alpha_4$ (Frequency). For small instances ($n=100$), the distribution shows a negative skew, indicating a preference for higher values. However, for large instances ($n=5000$), the strategy inverts, showing a positive skew and concentrating the probability mass on lower values. This sophisticated, scale-dependent adaptation is not observed in the other models, whose strategies remain largely fixed. We hypothesize that this ability to discern different patterns at different problem scales is a key reason for Llama-4-Maverick's superior performance.

\paragraph{The Impact of Instance Scale on Strategic Certainty.}
A general trend observed across all models is the influence of instance size on strategic ``certainty.'' For smaller instances ($n=100$), the violin plots are generally wider and more dispersed, suggesting that the LLMs are less certain in their assignments when provided with a smaller, less informative dataset. Conversely, for large instances ($n=5000$), the distributions become tighter and more defined. This indicates that all models are able to form more confident and consistent heuristic strategies when analyzing the richer data from larger problem instances.

\subsubsection{Analysis of Beta Parameter Distributions}

Unlike the alpha parameters, which represent interdependent weights, the beta parameters ($\beta_i$) are independent coefficients. They define the \textit{target or ideal value} that, according to the LLM, each corresponding metric should have for a run to be considered highly desirable.

\begin{figure}[t]
    \centering
    \includegraphics[width=0.9\linewidth]{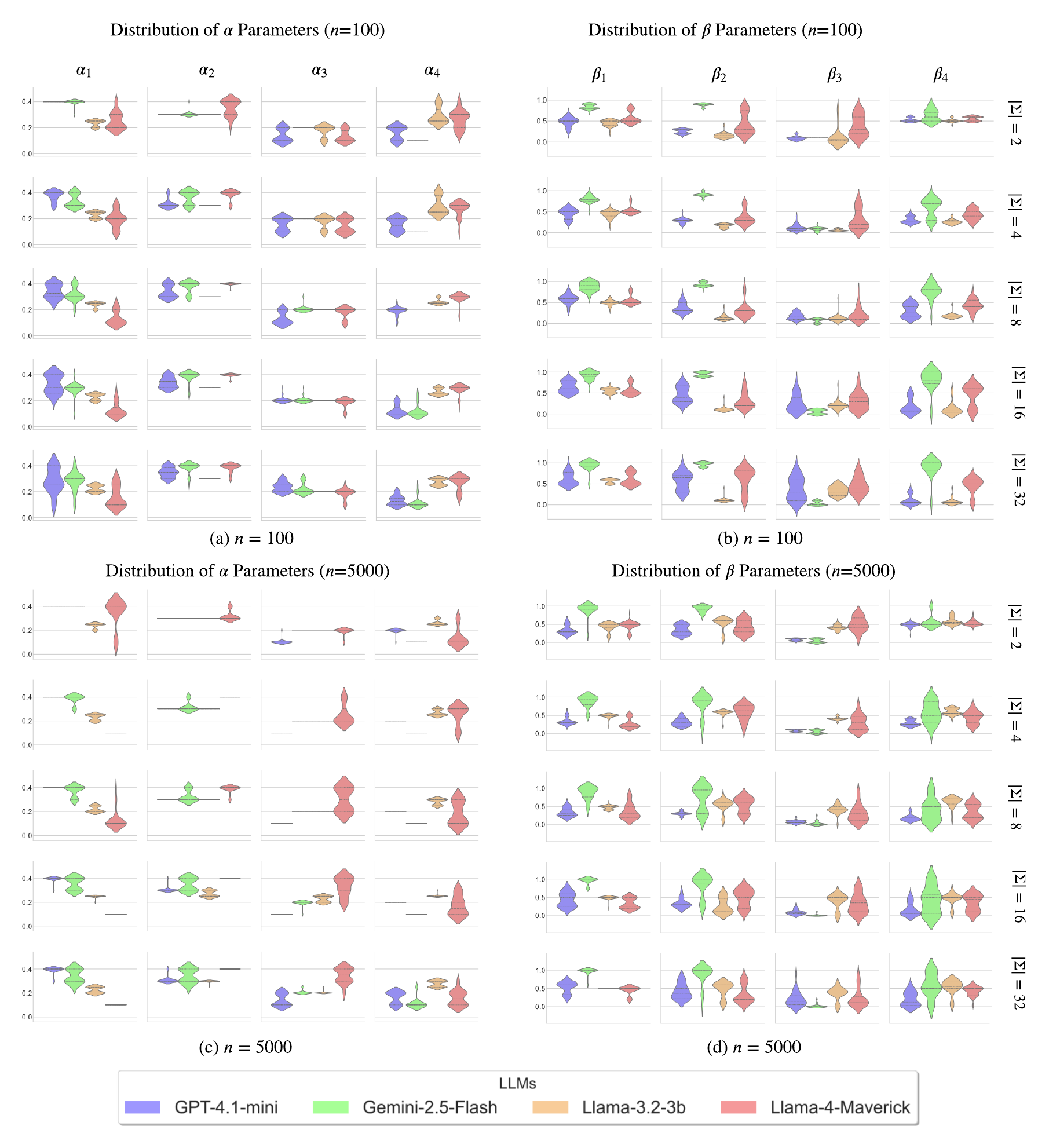}
    
   \caption{Violin plots showing the distribution of estimated parameters by four LLMs. The analysis compares parameter $\alpha$ (left) and $\beta$ (right) for input strings of length 100 (top) and 5000 (bottom). Each facet within the subplots corresponds to a different alphabet size $|\Sigma| \in \{2, 4, 8, 16, 32\}$.}
    \label{fig:alpha_beta_analysis}
\end{figure}

Figure~\ref{fig:alpha_beta_analysis}---subplots (b) and (d) in the second column---presents the findings for the $\beta_i$ parameters, to which the analysis now turns.

\paragraph{Consensus on Ideal Values: The Role of Length and Distance.}

The analysis reveals a strong consensus on the ideal values for two of the four metrics. For \textbf{$\beta_1$} (Normalized Length), all models consistently favor high values, confirming that longer runs are universally seen as desirable. An even stronger consensus exists for \textbf{$\beta_3$} (Distance to Next Run), where all models unanimously assign very low ideal values. This indicates a shared, fundamental strategy to prioritize runs that are part of tightly clustered, fragmented super-runs.

\paragraph{Divergent Strategies for Secondary Metrics.}

In contrast, significant strategic divergence is observed for \textbf{$\beta_2$} (Opportunity) and \textbf{$\beta_4$} (Global Character Frequency). For $\beta_2$, models like Llama-4-Maverick and Gemini-2.5-Flash are confident that the ideal value is high (near 1.0), whereas GPT-4.1-mini and Llama-3.2-3b exhibit much wider, more uncertain distributions. This suggests a split in strategy, where some models see high opportunity as a critical trait, while others are more uncertain. The most notable divergence occurs in $\beta_4$, where Gemini-2.5-Flash is the only model to consistently favor high character frequency, treating it as a desirable attribute that all other models consider neutral.

\paragraph{The Impact of Instance Scale on Strategic Confidence.}

A surprising trend emerges when comparing the distributions across different instance scales. Unlike the alpha parameters, where LLM certainty generally increased with larger data (at $n=5000$), the beta distributions show the opposite effect. For larger and more complex instances, the violin plots for beta values often become wider and more dispersed. We hypothesize that as the combinatorial noise increases, identifying clear numerical patterns that point to a single ``ideal'' value becomes more difficult. Consequently, the LLMs express this uncertainty by providing a broader distribution of what could be considered a ``good'' value. This is exemplified by Gemini's behavior on $\beta_2$ for $n=5000$: its distribution becomes highly skewed, suggesting it either detects complex instance-specific subtleties or its internal bias struggles to maintain consistency under high complexity.

\subsubsection{Synergy Between Alpha and Beta Parameters}

A holistic analysis of both the alpha and beta distributions reveals a consistent and sophisticated reasoning process across the different LLMs. We observe a shared tendency among the models to create a balanced heuristic profile: some metrics are assigned high importance (a high $\alpha$) and a high target value (a high $\beta$), while others are deliberately favored with low target values (a low $\alpha$ or low $\beta$). This indicates a nuanced, non-random strategy. Furthermore, the higher dispersion in some distributions (wider violin plots) suggests a sophisticated trade-off, where an LLM's uncertainty about one metric may be compensated for by a stronger conviction in another. This trade-off mechanism appears to be more refined in superior models like Llama-4-Maverick, whose ability to detect subtle numerical patterns likely explains the unique shapes of its parameter distributions (e.g., for the alpha parameters at $n=5000$). This aligns with a key insight from the foundational work by Sartori et al., who state that:
\begin{quote}
    ``The beta values play a crucial role in shaping the response quality. [...] By assigning importance weights to each metric (alpha values) and requesting an expected value (beta), we apparently enable the LLM to uncover more subtle patterns [...] ultimately leading to enhanced results.''~\cite[pag. 15-16]{10818476}
\end{quote}

Our results confirm that the alpha and beta parameters work synergistically to produce a robust heuristic for string problems like the LRS, with the absence of either parameter set significantly degrading performance. This synergistic effect likely arises because the LLM must reconcile two linguistically grounded yet conceptually distinct factors---\textit{importance} (alpha) and \textit{ideal} (beta)---which operate along different dimensions and do not naturally align.

\begin{table}
\centering
\sisetup{group-separator={,}} 

\caption{Summary of dataset token usage and estimated API processing costs for different LLMs.}
\label{tab:token_cost_analysis}

\begin{subtable}{\textwidth}
    \centering
    \caption{Token usage across the entire dataset, based on 150 instances per input string length}

    \label{tab:token_usage}
    \begin{tabular}{
        S[table-format=4.0]
        S[table-format=5.0]
        S[table-format=8.0]
    }
        \toprule
        {\textbf{Instance Length}} & {\textbf{Tokens per Instance}} & {\textbf{Total Tokens (150 Instances)}} \\
        \midrule
        100   & 1920   & 288000 \\
        200   & 2980   & 447000 \\
        300   & 4520   & 678000 \\
        500   & 8360   & 1254000 \\
        1000  & 15210  & 2281500 \\
        2000  & 33721  & 5058150 \\
        5000  & 90237  & 13535550 \\
        \midrule
        \multicolumn{2}{r}{\textbf{Total Tokens for Dataset}} & \textbf{23542200} \\
        \multicolumn{2}{r}{\textbf{Total Tokens (in Millions)}} & \textbf{23.54} \\
        \bottomrule
    \end{tabular}
\end{subtable}

\vspace{1cm} 

\begin{subtable}{\textwidth}
    \centering
    \caption{Estimated API cost based on OpenRouter pricing. *Llama-3.2-3b was available at no cost at the time of testing. EUR values are illustrative (1 USD = 0.93 EUR).}
    \label{tab:cost_estimation_eur}
    \begin{tabular}{
        l
        S[table-format=2.2]
        S[table-format=2.2]
        S[table-format=2.2]
        S[table-format=2.2]
    }
        \toprule
        {\textbf{Metric}} & {\textbf{GPT-4.1-mini}} & {\textbf{Gemini-2.5-Flash}} & {\textbf{Llama-3.2-3b}} & {\textbf{Llama-4-Maverick}} \\
        \midrule
        Input Cost (\$/1M tokens)  & 0.40   & 0.15  & {\text{Free*}} & 0.16 \\
        \textit{Input Cost (€/1M tokens)} & \textit{0.37} & \textit{0.14} & {\textit{---}}   & \textit{0.15} \\
        \addlinespace
        Output Cost (\$/1M tokens) & 1.60   & 0.60  & {\text{Free*}} & 0.60 \\
        \textit{Output Cost (€/1M tokens)} & \textit{1.49} & \textit{0.56} & {\textit{---}}   & \textit{0.56} \\
        \midrule
        \textbf{Estimated Total Cost (\$)} & \textbf{11.02} & \textbf{4.13} & \textbf{0.00}    & \textbf{4.37} \\
        \textbf{\textit{Estimated Total Cost (€)}} & \textbf{\textit{10.25}} & \textbf{\textit{3.84}} & \textbf{\textit{0.00}}    & \textbf{\textit{4.06}} \\
        \bottomrule
    \end{tabular}
\end{subtable}

\end{table}
\subsection{Analysis of Costs and API Latency}\label{sec:analysis-api-latency}

\paragraph{Token Consumption.} Using LLMs, even open-weight models, incurs both computational and financial costs. It is therefore important to analyze the practical implications of our framework. Table~\ref{tab:token_usage} summarizes the total token consumption for the 1050 LRS instances used in our experiments. A key finding from this data is that token consumption does not scale linearly with the instance length; instead, it exhibits a super-linear growth, as demonstrated by the following observations:
\begin{itemize}
    \item \textbf{Absolute Growth:} The absolute token increase between consecutive length benchmarks accelerates, from small increments ($\approx$1k tokens) for short input strings to a very large jump ($\approx$56k tokens) for the longest ones.

    \item \textbf{Relative Growth:} The relative (percentage) growth also increases with instance size, peaking at a +168\% increase between input strings of length 2000 and 5000.

    \item \textbf{Tokens per Unit of Length:} The token-to-length ratio is not constant, oscillating between approximately 15 and 19, with a tendency to increase for very large instances.
\end{itemize}
These observations collectively suggest that the token cost grows super-linearly with instance length.

\paragraph{Financial Cost Analysis.} Table~\ref{tab:cost_estimation_eur} translates this token usage into estimated financial costs based on OpenRouter's pricing. The analysis reveals a critical trade-off between cost and performance. GPT-4.1-mini emerges as the most expensive model, a significant finding given that it was also one of the lower-performing LLM-hybrids. This underscores that the choice of LLM is not trivial, as a suboptimal model can lead to excessive costs without practical advantages in solution quality. Conversely, Llama-3.2-3b, being a free-to-use model on the platform, presents itself as a viable alternative for scenarios with limited financial resources. However, as we will discuss next, this does not equate to zero-cost when considering computational time.

\begin{figure}
\begin{tikzpicture}
\begin{axis}[
    title={LLM API Latency},
    width=\linewidth,
    height=8cm,
    ybar,
    bar width=0.3cm,
    enlarge x limits=0.1,
    xlabel={Length instances},
    ylabel={Time (seconds)},
    symbolic x coords={100,200,300,500,1000,2000,5000},
    xtick=data,
    ymin=0,
    legend style={
        at={(0.5,-0.25)}, 
        anchor=north, 
        legend columns=4, 
        column sep=10pt
    },
    tick label style={font=\small},
    label style={font=\small},
    title style={font=\small}
]
\addplot[fill=blue!50] coordinates {(100,2.138) (200,1.896) (300,2.009) (500,1.882) (1000,3.498) (2000,3.057) (5000,7.282)};
\addlegendentry{GPT-4.1-mini}

\addplot[fill=green!50] coordinates {(100,2.152) (200,2.116) (300,2.117) (500,2.315) (1000,2.709) (2000,1.995) (5000,10.499)};
\addlegendentry{Gemini-2.5-Flash}

\addplot[fill=orange!50] coordinates {(100,2.795) (200,3.253) (300,3.468) (500,4.865) (1000,9.398) (2000,14.604) (5000,21.316)};
\addlegendentry{Llama-3.2-3b}

\addplot[fill=red!50] coordinates {(100,2.056) (200,1.234) (300,1.623) (500,1.726) (1000,3.109) (2000,2.891) (5000,3.815)};
\addlegendentry{Llama-4-Maverick}

\node at (axis cs:100,2.056) [above=1pt, xshift=1.125*0.5cm] {\scriptsize\textcolor{orange}{\faBolt}};
\node at (axis cs:200,1.234) [above=1pt, xshift=1.125*0.5cm] {\scriptsize\textcolor{orange}{\faBolt}};
\node at (axis cs:300,1.623) [above=1pt, xshift=1.125*0.5cm] {\scriptsize\textcolor{orange}{\faBolt}};
\node at (axis cs:500,1.726) [above=1pt, xshift=1.125*0.5cm] {\scriptsize\textcolor{orange}{\faBolt}};
\node at (axis cs:1000,2.709) [above=1pt, xshift=-5] {\scriptsize\textcolor{orange}{\faBolt}};
\node at (axis cs:2000,1.995) [above=1pt, xshift=-5] {\scriptsize\textcolor{orange}{\faBolt}};
\node at (axis cs:5000,3.815) [above=1pt, xshift=1.125*0.5cm] {\scriptsize\textcolor{orange}{\faBolt}};

\end{axis}
\end{tikzpicture}
\caption{API latency (seconds) comparison for four LLM variants---GPT-4.1-mini, Gemini-2.5-Flash, Llama-3.2-3b, and Llama-4-Maverick---across various input lengths using OpenRouter. A lightning bolt icon (\textcolor{orange}{\faBolt}) highlights the fastest model in each group.}
\label{fig:llm_time_comparison_en_annotated}
\end{figure}
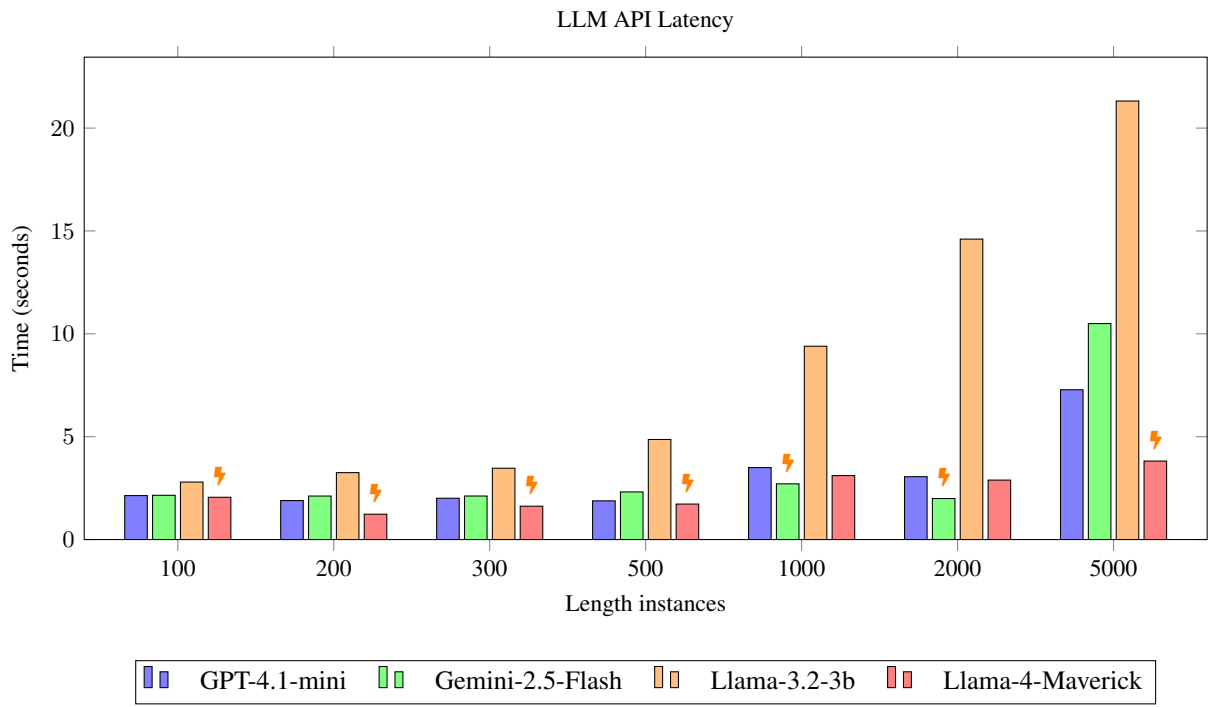

\paragraph{API Latency.}
Beyond financial cost, API latency is a critical factor for the framework’s practical deployment, particularly since the LLMs are accessed via external infrastructure through a proxy service such as OpenRouter. Figure~\ref{fig:llm_time_comparison_en_annotated} presents the response time of each LLM as a function of instance length. The results reveal distinct performance profiles: Llama-4-Maverick demonstrates the most stable behavior, maintaining low and consistent latency across all instance sizes. It is generally the fastest model, although Gemini-2.5-Flash is marginally quicker for mid-sized instances. In contrast, the free-to-use Llama-3.2-3b shows a pronounced increase in latency as instance length grows, indicating a trade-off in which the absence of financial cost is compensated by lower processing priority on the host infrastructure.

Notably, Llama-4-Maverick not only delivers strong performance, as demonstrated throughout this work, but also consistently responds quickly---an aspect that should be taken into account when response time is a critical requirement.

\subsubsection{Practical Considerations}

Our results offer several practical considerations for researchers looking to implement similar frameworks. First, regarding model selection, our findings suggest that beginning with a free-to-use, open-weight model like Llama-3.2-3b is a prudent strategy. Depending on the initial results, one can then scale up to larger models if necessary, keeping in mind that a higher cost does not guarantee better performance, as demonstrated by the case of GPT-4.1-mini. Second, we recommend using a proxy service like OpenRouter rather than committing to a single model's API. This approach provides the flexibility to benchmark multiple models and identify the most suitable one for a given problem without significant engineering overhead.

\section{Discussion}\label{sec:dis}
\paragraph{Interpretation of the Instance-Driven Heuristic Approach.}

Our results indicate that instance-driven heuristic generation via metric analysis is a highly promising approach. As discussed in Section~\ref{sec:relation_ec}, it occupies a distinct space relative to traditional hyper-heuristics and dynamic parameter adaptation. The strong performance of our method, particularly in the context of complex instances with large alphabet sizes, suggests that this form of a priori guidance is especially valuable when the search landscape is noisy and combinatorial complexity is high. In such scenarios, providing the BRKGA with a robust initial bias allows it to avoid suboptimal regions where purely dynamic or selection-based methods may struggle at the outset.

A key novelty of this framework is the redefinition of the algorithm designer’s role. Rather than manually constructing a portfolio of low-level heuristics for use in a hyper-heuristic framework, the designer now operates at a higher level of abstraction: shaping the collaborative process and crafting the prompts that elicit useful analytical behavior from the LLM. As our ablation study on metric selection demonstrates, this human–LLM collaboration is crucial not only for producing high-quality heuristics but also for ensuring computational feasibility---a nuance that differentiates our approach from purely automated discovery methods.

BRKGA proves particularly well-suited to this framework because its search dynamics can be strongly influenced by small, well-designed biases. This advantage is most evident in the most challenging LRS instances, where BRKGA+LLM consistently outperformed the baseline in statistical tests (see~\cref{tab:statistical_summary_final,tab:wilcoxon_significant}) and, as shown in the STNWeb analysis (Figure~\ref{fig:stnweb-analysis}), avoided the premature convergence observed in the pure BRKGA.

Finally, our ablation study underscores that such guidance must be coherent. When the bias vector is derived from random values rather than meaningful patterns, it becomes a negative bias---slowing convergence and increasing the risk of stagnation.

\paragraph{The Role of Human-LLM Collaboration.}

A key distinction of our framework is its emphasis on strengthening human–LLM collaboration rather than replacing the human designer. Instead of using the LLM to automate every facet of the research process, we harness its core strength---detecting subtle patterns in complex data---to augment human expertise. This human-in-the-loop methodology ensures that final design choices are not only intelligently guided but also rigorously validated.

One might argue that a simpler supervised model (e.g., logistic regression) could learn weights from the selected metrics. Yet such methods demand labeled solutions, retraining for each domain, and manual feature pipelines. By contrast, our LLM-based approach works in a zero-shot manner---no data, no retraining, and no extra code beyond generic prompts---enabling immediate use across heterogeneous problems. Even if inference cost is higher, development cost is drastically reduced. We thus position LLMs not as lighter ML alternatives but as flexible heuristic biases.

This collaboration proved essential not only for heuristic quality but also for practical feasibility. As revealed by our ablation study, LLMs were able to propose highly effective metrics that, however, were computationally costly. Here, the expert’s role was crucial: filtering these suggestions to produce a final set of metrics that remained computationally efficient---an algorithmic complexity nuance the LLM alone could not discern.

\paragraph{Novel Contributions.}

Our results reaffirm and extend the conclusions of the foundational work~\cite{10818476}. We confirm that BRKGA is a strong candidate for the instance-driven approach, although whether it is the optimal choice remains an open question. This provides strong evidence that the approach is robust and potentially applicable to a wide range of combinatorial optimization problems. Importantly, we show that the method is effective not only with large proprietary LLMs, but also with open-weight and smaller language models (SLMs).

We significantly broaden the scope of the original framework in several ways. While the foundational study focused on a social network problem, we successfully adapted the approach to a completely different domain: string-based problems in bioinformatics. We also expanded the LLM’s role to include cooperation in the metric selection phase, and conducted an exhaustive behavioral analysis of the method---examining, for instance, how the alpha–beta parameters respond to instance characteristics---an aspect absent from the original work.

\paragraph{Limitations and Methodological Insights.}

A key methodological insight from our work concerns the specific architectural properties that make a metaheuristic receptive to this instance-driven framework. Our findings confirm that BRKGA is an exceptionally suitable candidate, but they do not support the claim that this framework is universally applicable. This distinction provides a roadmap for future research in this area.

The suitability of BRKGA stems directly from its core design: the separation between a problem-agnostic encoding and a problem-specific decoder. A solution is encoded as a vector of random keys, which the decoder interprets as a prioritized construction order. The multiplicative bias ($\vec{v}_i \cdot \vec{L}_i$) proposed in our framework integrates seamlessly into this process (Algorithm~\ref{alg:brkga-llm-bias}). It directly modulates the priority of each solution component in an intuitive manner, effectively steering the greedy construction without destroying the underlying stochastic exploration provided by the genetic operators. This clean decoupling allows the external, static guidance from the LLM to be injected precisely at the interface between the search and construction phases.

However, this compatibility is not a property of all metaheuristics. Many prominent algorithms do not rely on a single, prioritized construction list. Instead, their search may be guided by emergent, collective memory structures that consolidate over time, or by the interaction of solutions within a continuous search space~\cite{10.1145/937503.937505}. In such cases, a simple, static bias applied at the start is unlikely to exert a lasting influence, as it may be quickly dominated by the algorithm's own dynamic learning and exploration mechanisms.

Consequently, adapting the LLM-generated guidance to these other strategies would require more deeply integrated and novel mechanisms beyond the multiplicative approach used here. A structural modification of the core heuristic---such as influencing a reinforcement learning update rule or a particle's velocity calculation---might be necessary, rather than a simple pre-computation bias. We speculate that the success of this framework is therefore tightly coupled to the internal search mechanism of the chosen metaheuristic.

\section{Conclusion}\label{sec:conclusion}

In this work, we demonstrated that the instance-driven heuristic bias paradigm for integrating metaheuristics with Large Language Models (LLMs) can be successfully adapted to tackle string-based problems, such as the Longest Run Subsequence. We proposed an extension of this paradigm by introducing a human-LLM collaborative process for co-designing and implementing a feature set, as well as generating heuristic guidance, which was subsequently incorporated into a Biased Random-Key Genetic Algorithm (BRKGA).

Our comprehensive experimental evaluation---including ablation studies, behavioral analyses, and rigorous statistical tests---validated the effectiveness of this framework. The results indicate that our best-performing hybrid, BRKGA+Llama-4-Maverick, achieved a statistically significant improvement over the baseline in 42.86\% of instance groups, with the greatest advantage observed on the most complex instances. These findings confirm that leveraging an LLM as an instance-driven pattern-recognition engine is a promising and effective approach to enhancing metaheuristics in complex combinatorial domains.

Future work should explore how this framework can be extended to other types of metaheuristics and applied to different families of combinatorial optimization problems. Moreover, given that this approach  is modular rather than closed, future research could explore how it can be integrated into established frameworks that combine LLMs with metaheuristics.

\section*{Acknowledgments}

C.~Chacón~Sartori and C.~Blum were supported by grant PID2022-136787NB-I00 funded by MCIN/AEI/10.13039/501100011033. P. Pinacho-Davidson acknowledges financial support from FONDECYT through grant number 11230359.

\bibliographystyle{plainnat}  
\bibliography{references}

\end{document}